\documentclass{article}
\usepackage{graphicx} 
\usepackage{arxiv}
\usepackage[utf8]{inputenc} %
\usepackage[T1]{fontenc}    %
\usepackage{hyperref}       %
\usepackage{url}            %
\usepackage{booktabs}       %
\usepackage{amsmath}       %
\usepackage{amsfonts}       %
\usepackage{nicefrac}       %
\usepackage{microtype}      %
\usepackage{cleveref}       %
\usepackage{natbib}
\usepackage{doi}
\usepackage{balance}
\usepackage{xcolor}
\usepackage{mathtools}
\usepackage{algorithm}
\usepackage{algpseudocode}
\usepackage{wrapfig}
\usepackage{subcaption}
\usepackage{xspace}

\usepackage{multicol}
\usepackage{tabularx}

\newcommand{\SD}{SD\xspace}  

\newcommand{\SAttn}{SA-Cross\xspace}
\newcommand{\TAttnSelf}{TA-Self\xspace}
\newcommand{\TAttnCross}{TA-Cross\xspace}

\definecolor{purple}{rgb}{0.3, 0, 1.0}
\definecolor{darkred}{rgb}{0.6, 0.1, 0.1}

{}

{}  

{}

\newcommand{\TitleLong}{TrailBlazer\xspace}  
\newcommand{\thetitle}{\TitleLong: Trajectory Control for Diffusion-Based Video Generation} 

{}  
\newcommand\KMtext[1]{{#1}}

\newcommand{\FIXLATER}[1]{}  
\newcommand{\LONGVERSION}[1]{}  

\newcommand{\SecRef}[1]{{Sec.~\ref{#1}}}   
\newcommand{\FigRef}[1]{{Fig.~\ref{#1}}}   
\newcommand{\EqRef}[1]{{Eq.~\ref{#1}}}
\newcommand{\link}[1]{\href{#1}{[link]}}

\newcommand{\bb}{\mathbf{b}}

\newcommand{\mm}{\mathbf{m}}

\newcommand{\zz}{\mathbf{z}}

\newcommand{\bAA}{\mathbf{A}}
\newcommand{\BB}{\mathbf{B}}
\newcommand{\CC}{\mathbf{C}}

\newcommand{\KK}{\mathbf{K}}

\newcommand{\MM}{\mathbf{M}}

\newcommand{\QQ}{\mathbf{Q}}

\newcommand{\VV}{\mathbf{V}}

\newcommand{\BBB}{\mathcal{B}}

\newcommand{\III}{\mathcal{I}}

\newcommand{\KKK}{\mathcal{K}}

\newcommand{\PPP}{\mathcal{P}}

\newcommand{\RRR}{\mathcal{R}}

\newcommand{\TTT}{\mathcal{T}}

\newcommand{\Reals}{\mathbb{R}}	

\graphicspath{{assets/}}     

\title{\TitleLong: Trajectory Control for Diffusion-Based Video Generation}

\author{
    \href{https://www.linkedin.com/in/kurt-ma/}{Wan-Duo Kurt Ma}\\
    Victoria University of Wellington\\
    \texttt{mawand@ecs.vuw.ac.nz}\\
\And
    \href{http://scribblethink.org/}{J. P.~Lewis}\\  %
    NVIDIA Research\\
    \texttt{jpl@nvidia.com}\\
\And
    \href{https://people.wgtn.ac.nz/bastiaan.kleijn}{W. Bastiaan Kleijn}\\  %
    Victoria University of Wellington\\
    bastiaan.kleijn@vuw.ac.nz
}

\begin{document}
\vspace{-0.1in}
\maketitle
\vspace{-0.6in}

\begin{center}
    \centering \captionsetup{type=figure}
    \includegraphics[width=\textwidth]{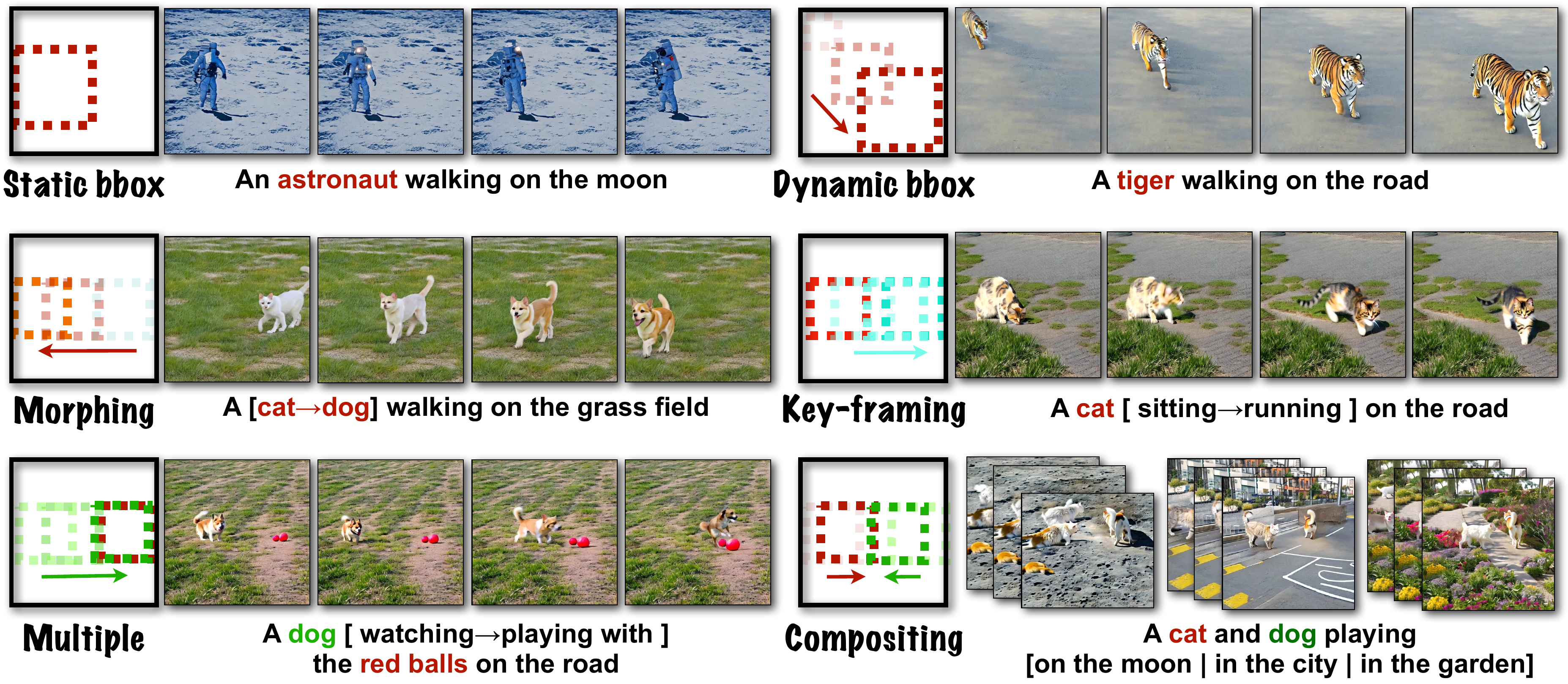}
    \captionof{figure}{\emph{\TitleLong} extends a pre-trained video diffusion model to introduce trajectory control over one or multiple subjects. Its primary contribution lies in the ability to animate the synthesized subject using a bounding box \emph{(bbox),} whether it remains static (Top-left) or dynamic in terms of location and bbox size (Top-right), morphing for subject interpolation (Middle-left), and varied movement speed (Middle-right). The moving subjects fit naturally within an environment specified by the overall prompt (Bottom-right). Additionally, the speed of the subjects can be controlled through keyframing (Bottom-left).}
    \label{fig:teaser}
\end{center}


\begin{abstract}
    Within recent approaches to text-to-video (T2V) generation, achieving controllability in the synthesized video is often a challenge. Typically, this issue is addressed by providing low-level per-frame guidance in the form of edge maps, depth maps, or an existing video to be altered. However, the process of obtaining such guidance can be labor-intensive. This paper focuses on enhancing controllability in video synthesis by employing straightforward bounding boxes to guide the subject in various ways, all without the need for neural network training, finetuning, optimization at inference time,  or the use of pre-existing videos. Our algorithm, \emph{\TitleLong}, is constructed upon a pre-trained (T2V) model, and easy to implement. \footnote{Our project page: \href{https://hohonu-vicml.github.io/Trailblazer.Page/}{https://hohonu-vicml.github.io/Trailblazer.Page/}} The subject is directed by a bounding box through the proposed spatial and temporal attention map editing. Moreover, we introduce the concept of keyframing, allowing the subject trajectory, morphing, and overall appearance to be guided by \emph{both} a moving bounding box and corresponding prompts, without the need to provide a detailed mask. The method is efficient, with negligible additional computation relative to the underlying pre-trained model. Despite the simplicity of the bounding box guidance, the resulting motion is surprisingly natural, with emergent effects including perspective and movement toward the virtual camera as the box size increases.
\end{abstract}


\section{Introduction}
\label{sec:intro}

Advancements in generative models for text-to-image (T2I) have been dramatic \cite{DALLE2,IMAGEN,RombachStablediffusion21,ediffI}. Recently, text-to-video (T2V) systems have made significant strides, enabling the automatic generation of videos based on textual prompt descriptions \cite{Ho2022ImagenVH, ho2022video, wu2023tune, Esser2023StructureAC}. One primary challenge in video synthesis lies in the extensive memory and required training data. Methods based on the pre-trained Stable Diffusion (SD) model have been proposed to address the efficiency issues in T2V synthesis. These approaches address the problem from several perspectives including finetuning  and zero-shot learning \cite{text2video-zero, qi2023fatezero}.

However, text prompts do not provide good control over the spatial layout and trajectories of objects in the generated video. This control is known to be required for understandable narration of a story \cite{FilmGrammar}. Existing work such as \cite{hu2023videocontrolnet} has approached this problem by providing low-level control signals, e.g.,~using Canny edge maps or tracked skeletons to guide the objects in the video using ControlNet \cite{controlnet}. These methods achieve good controllability, but they can require considerable effort to produce the control signal. For example, capturing the desired motion of an animal (e.g.,~a tiger) or an expensive object (e.g.,~a jet plane) would be quite difficult, while sketching the desired movement on a frame-by-frame basis would be tedious.

To address the needs of casual users, we introduce a high-level interface for the control of object trajectories in synthesized videos summarized in \FigRef{fig:teaser}. Users simply provide bounding boxes (bboxes) specifying the desired position of an object at several times (keyframes) in the video, together with the text prompt(s) describing the object at the corresponding times. The provided bboxes are interpolated between the keyframes, resulting in smooth motion and size changes of the object. For instance, the cat sitting in the early half of the video in the red bbox, and then moving with the cyan bbox achieved through keyframing in the middle right of \FigRef{fig:teaser}.

If more than one different text prompt is provided, these prompts are also interpolated, resulting in a ``morphing'' effect such as the cat that transforms into a dog in \FigRef{fig:teaser}. To achieve this guidance we take inspiration from the observation \cite{Magicmix} that object position is established early in the denoising diffusion process, and we leverage the clear spatial interpretation of spatial and temporal attention maps as illustrated in \FigRef{fig:mot}.
Our resulting strategy involves editing \emph{both spatial and temporal attention maps} for a specific object during the initial denoising diffusion steps to concentrate activation at the desired object location. Our inference-time editing approach achieves this without disrupting the learned text-image association in the pre-trained model, and requires minimal code modifications. 

Our method, \textit{TrailBlazer}, builds on previous works. We use the pre-trained ZeroScope model \cite{zeroscope}, which is a fine-tuned version of \cite{modelscope}, as our underlying model. A body of previous and concurrent works have addressed guiding object position in image generation models, including \cite{layout2im,L2Igan,L2Ixformer,ediffI,ma2023directed,boxdiff,GLIGEN,multidiffusion}.
TrailBlazer most closely resembles the cross-attention injection used in \cite{ma2023directed}, and we adopt some notation from that paper. However, our work addresses a different problem, that of controlling position and trajectories in \emph{videos}, which requires a different approach to control temporal cross-frame attention. Our work also does not require the inference-time optimization algorithm used in \cite{ma2023directed}.

Our contributions are three-fold:

\begin{itemize}
    \item \textbf{Novelty.} We introduce a novel approach employing high-level bounding boxes to guide the subject in diffusion-based video synthesis. This approach is suitable for casual users, as it avoids the need to record or draw a frame-by-frame positioning control signal. In contrast, the low-level guidance signals such as detailed masks, edge maps, used by some other approaches have two disadvantages: it is difficult for non-artists to draw these shapes, and processing existing videos to obtain these signals limits the available motion to copies of existing sources.

    \item \textbf{Position, size, and prompt trajectory control.} Our approach enables users to position the subject by keyframing its bounding box. The size of the bbox can be similarly controlled, thereby producing perspective effects (Figs.~\ref{fig:teaser},\ref{fig:main_dyn}).
    Finally, users can also keyframe the text prompt to
    influence the behavior and identity of the subject in the synthesized video (Figs.~\ref{fig:teaser}).

    \item \textbf{Simplicity.} Our method operates by directly editing the spatial and temporal attention in the pre-trained denoising UNet. It requires no training or optimization, and the core algorithm can be implemented in less than 200 lines of code.

\end{itemize}
\section{Related Work}
\label{sec:related}

\subsection{Text-to-Image (T2I)}

Denoising diffusion models construct a stochastic \cite{sohldickstein15,songScorematching19,hoDDPM20} or deterministic \cite{DDIM} mapping between the data space and a corresponding-dimension multivariate Gaussian. Signals are synthesized by sampling from a normal distribution and performing a sequence of denoising steps. A number of works \cite{GLIDE,nichol2021improved,DALLE2,Saharia2022PhotorealisticTD} have performed T2I synthesis using images conditioned on the text embedding from a model such as CLIP \cite{CLIP}. Performance is significantly improved in the Latent Diffusion Model \cite{RombachStablediffusion21} (LDM) by performing the diffusion computation in the latent space of a carefully trained variational autoencoder. LDM was trained with a large scale dataset, resulting in the widely adopted Stable Diffusion (\SD) system. We omit the basic diffusion derivation as tutorials are available, e.g., \cite{blogLillog}.

Despite the success of image generation using SD, it is widely acknowledged that SD lacks controllability in synthesis. SD faces challenges in synthesizing multiple objects, often resulting in missing objects or incorrect assignment of prompt attributes to different objects. Recently, ControlNet \cite{controlnet} and T2I-Adapter \cite{mou2023t2i} introduced additional fine-tuning layers to train the model with various forms of image conditioning such as edge maps, or rigging skeletons.

The methods of \cite{layout2im,L2Igan,L2Ixformer,ma2023directed,boxdiff,multidiffusion} have addressed the layout-to-image (L2I) issue using few-shot learning. Directed Diffusion \cite{ma2023directed}, BoxDiff \cite{boxdiff}, and MultiDiffusion \cite{multidiffusion} use coarse bboxes to control subject position, achieving good results by manipulating the spatial latent and text embeddings cross attention map \cite{PromptToPrompt}.

\subsection{Text-to-Video (T2V)}

Text-to-video (T2V) synthesis is generally more difficult than T2I due to the difficulty of ensuring temporal consistency and the requirement for a large paired text and video dataset. \cite{ho2022video,harvey2022flexible,Infilling2022,voleti2022MCVD,yang2022diffusion,nvidianoiseprior} show methods that build on top of image diffusion models. Some works \cite{blattmann2023videoldm,luo2023videofusion} also introduce 3D convolutional layers in the denoising UNet to learn temporal information. Imagen Video \cite{Ho2022ImagenVH} achieves higher resolution by computing temporal and spatial super-resolution on initial low resolution videos. VideoLDM \cite{blattmann2023videoldm} and ModelScope \cite{luo2023videofusion} insert a temporal attention layer by reshaping the latent tensor. Text2Video-Zero \cite{text2video-zero}, \KMtext{denoted as T2V-Zero}, and FateZero \cite{qi2023fatezero} investigate how the temporal coherence can be improved by cross frame attention manipulation with pre-trained T2I models. \cite{nvidianoiseprior} addresses the same problem by introducing temporal correlation in the diffusion noise.
However, these pioneering studies generally lack position control in the video synthesis.

Recently several works have been proposed to solve the controllability in video synthesis problem by using pre-trained models together with low-level conditioning information such as edge or depth maps. Control-A-Video \cite{chen2023controlavideo} and MagicProp \cite{yan2023magicprop} use depth maps with ControlNet to train a temporal-aware network. T2V-Zero \cite{text2video-zero} partially achieves controllability by initializing the latent frames conditioned on the first frame with applied linear translation. However, the control is indirect and requires two steps. First, the user first needs to locate the subject's numerical location, and then adjust a translation offset. Distinct from the methods above, we use an attention injection method to guide the denoising path rather than optimization, and in general this is robust to different random seeds. The recent project Peekaboo \cite{jain2023peekaboo} is concurrent with TrailBlazer and shares similar goals.  Both Peekaboo and TrailBlazer guide subjects in video by manipulating the attention, however the formulations differ in many details. Peekaboo's use of an infinite negative attention injection in the background regions appears to often result in backgrounds with missing detail. 
In \SecRef{sec:exp}, we will provide both quantitative and visual evidence to demonstrate the better controllability and quality of our results. 
Other very recent preprints that address the layout-to-video problem in differing ways include \cite{lian2023llmgrounded, wang2024boximator, yang2024directavideo}. We do not compare against these concurrent works because their source was not available at the time of writing.

\section{Method}

\label{sec:method}

TrailBlazer is based on the open-source pre-trained model ZeroScope \cite{zeroscope}. This is a fine-tuned version of ModelScope \cite{luo2023videofusion}, known for its ability to generate high-quality videos without significant temporal flickering. It is noteworthy that TrailBlazer preserves this desirable temporal coherence effect achieved in their work.  TrailBlazer does not require any training, optimization, or low-level control signals (e.g., edge, depth maps with ControlNet \cite{controlnet}).  On the contrary, all that is required from the user is the prompt and an approximate bounding box (bbox) of the subject. Bboxes and corresponding prompts can be specified at several points in the video, and these are treated as \emph{keyframes} and interpolated to smoothly control both the motion and prompt content. 

We use the following notation: Bold capital letters (e.g., $\MM$) denote a matrix or a tensor depending on the context, vectors are represented with bold lowercase letters (e.g., $\mm$), and  scalars are denoted as lowercase letters (e.g., $m$). We use superscripts to denote an indexed tensor slice (e.g., $\MM^{(i)}$).  A synthesized video is composed of a number of images ordered in time. The individual images will be referred to as \emph{frames}, and the collection of corresponding times is the \emph{timeline}.
Spatial or temporal attention will be informally referred to as \emph{correlation}.

\begin{figure}[ht]
    \centering
    \includegraphics[width=\linewidth]{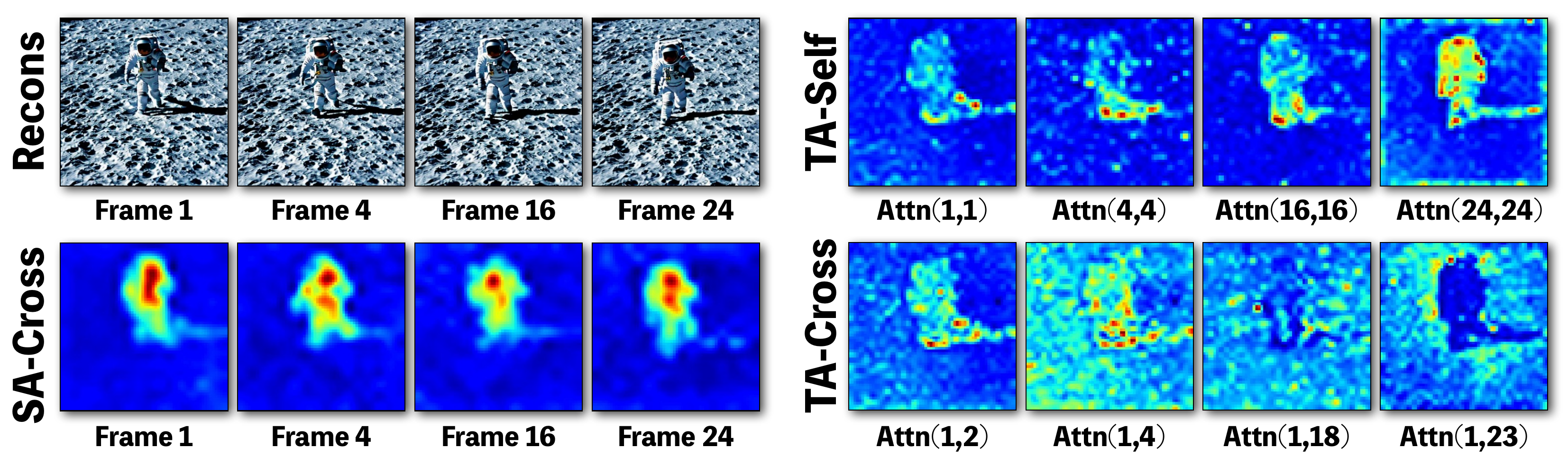}
    \caption{\textbf{Basis of our method.} We draw inspiration from inspection of the spatial (SA) and temporal (TA) attention maps viewed with self-frame attention (Self) and cross-frame attention (Cross). Thus, \TAttnSelf and \TAttnCross denote the self- and cross-frame attention map, respectively. \SAttn is the spatial cross-attention map with the prompt word ``astronaut''. The symbol ``$\text{Attn(i,j)}$'' denotes the temporal attention map between frame $i$ and frame $j$. The Recons subfigure shows reconstructions sampled from frames 1, 4, 16, and 24, respectively. In the \TAttnCross, the frame number were manually chosen to best illustrate the cross-frame attention between the astronaut and the background. Please refer to the main text for more details.}
    \label{fig:mot}
    \vspace{-1em}
\end{figure}

Similar to the work in \cite{ma2023directed}, our method draws significant inspiration from visual inspection of cross-attention maps. Consider
the final cross-attention result depicted in \FigRef{fig:mot}, generated from the prompt ``an astronaut walking on the moon''. The spatial cross attention, denoted as \SAttn, associated with the prompt word ``astronaut'' is highlighted at the left of the second row, showcasing the overall position of the subject. Furthermore, we visualize the attention map from the temporal module in the pre-trained model. The right in the first row displays ``self-frame'' temporal attention maps, denoted as \TAttnSelf, which consistently align with \SAttn.

The right in the second row of \FigRef{fig:mot} presents the visualization of cross-frame temporal attention maps, denoted as \TAttnCross, illustrating the attention between the first frame and subsequent frames in the video. As the distance between frames increases, the attention becomes less correlated in the subject area and becomes more correlated in the background area. This observation aligns with the reconstructed video shown in the left of the first row, where the background remains nearly static while the astronaut's position varies frame by frame. We will consider the temporal attention in detail in  \SecRef{ssec:tattn}.

\subsection{Pipeline}
\label{ssec:pipe}

\begin{figure*}
  \centering
  \includegraphics[width=\textwidth]{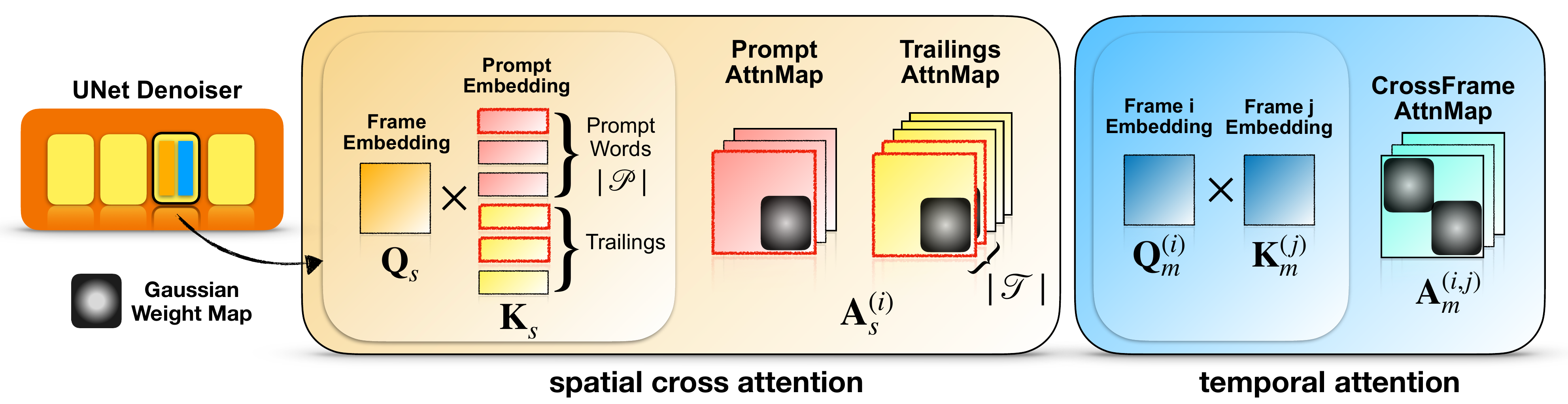}
  \caption{\textbf{Pipeline Overview.} Our pipeline highlights the central components of spatial cross-attention editing (left, in the blanched almond-colored section) and temporal cross-frame attention editing (right, in the blue section). This operation is exclusively applied during the denoising process in the early stage. The objective is to alter the attention map (e.g., $\bAA_s, \bAA_m$) using a Gaussian weighting within a user-specified bbox. This example uses one prompt word AttnMap and two trailing AttnMaps for guidance as highlighted in red.}
  \label{fig:pipe}
\end{figure*}

As mentioned above, keyframing \cite{keyframe} is a technique that defines properties of images at particular frames (keys) in a timeline and then automatically interpolates these values to achieve a smooth transition between the keys. It is widely used in the movie animation and visual effects industries since it reduces the artist's work while simultaneously producing temporally smooth motion that would be hard to achieve if the artist directly edited every image.
Our system takes 
advantage of this principle, and asks the user to specify several keys, consisting of bboxes and the associated prompts, describing the subject location and appearance or behavior at the particular times. For instance, as shown in \FigRef{fig:teaser} (Middle-right), the video of the cat initially sitting on the left, then running to the right, is achieved simply by placing keys at three frames only. Specifically, the sitting cat in the first part of the video is obtained with two identically positioned bboxes on the left, with the keyframes at the beginning and middle of the timeline and the prompt word ``sitting'' associated with both. A third keyframe is placed at the end of the video, with the bbox positioned on the right together with the prompt changing to ``running''. This results in the cat smoothly transitioning from sitting to running in the second part of the video.


We use the pre-trained ZeroScope model \cite{zeroscope} in all our experiments with no neural network training, finetuning, or optimization at inference time. Our pipeline is shown in \FigRef{fig:pipe}. The spatial cross attention and the temporal attention is discussed in detail in the \SecRef{ssec:sattn} and \SecRef{ssec:tattn}, respectively. 
All spatial and temporal editing is performed in the early steps $t \in \{T,...,T-N_S\}$, and $t \in \{T,...,T-N_M\}$ of the backward denoising process, where $T$ is the total number of denoising time steps, and $N_S$, and $N_M$ are hyperparameters specifying the number of steps of spatial and temporal attention editing. The parameter settings are detailed in our supplementary material.

In the subsequent sections we describe how our algorithm is implemented by modifying the spatial and temporal attention in a pre-trained diffusion model. Please refer to \cite{RombachStablediffusion21, DDIM, hoDDPM20, blogLillog} for background on overall diffusion model architectures.


Our system processes a set of keyframes, encompassing associated bbox regions $\RRR_f$ and prompts $\PPP_f$ at frame $f$, where $f$ denotes the frame index within the range $f \in \{1,...,N_F\}$. The users are required to specify a minimum of two keyframes: one at the start and one at the end of the video sequence.
The information in these keyframes is linearly interpolated, such as the bbox $\BBB_f$ and the prompt text embedding $y(\PPP_f)$ through the text encoder $y(\cdot)$. To enhance readability, we omit the subscript $f$ and the linearly blended video sequence between the keyframes when discussing the core method.

A region $\RRR$ is characterized by a set of parameters $\RRR = \{\BBB, \III, \TTT\}$: a set of bbox positions (e.g., $\BBB$), the indices of the subject we would like to constrain (e.g., $\III$), and the indices of the trailing maps (e.g., $\TTT$) to enhance the controllability. The subject indices $\III \subset \{ i | i \in \mathbb{N}, 1 \leq i \leq |\PPP| \}$, are 1-indexed with the associated word in the prompt. For example, $\III = \{1,2\}$ is associated with ``a'', ``cat'' in the prompt ``a cat sitting on the car''. 

The trailing attention maps indices $\TTT \subset \{ i | i \in \mathbb{N}, |\PPP| < i \leq N_P \}$ is the set of indices corresponding to the cross-attention maps generated without a prompt word association, where $N_P$ denotes the maximum prompt length that a tokenizer model can take, which is $N_P=77$ when CLIP is used \cite{CLIP}. The trailing attention maps serve as a means of controlling the spatial location of the synthesized subject and its attributes. A larger trailing indices set $|\mathcal{T}|$ provides greater controllability but comes with the risk of failed reconstruction \cite{ma2023directed}.

A bbox $\BBB = \bigl\{ (x, y) \,| \, b_{\text{left}} \times w \leq x \leq b_{\text{right}} \times w, \; b_{\text{top}} \times h \leq y \leq b_{\text{bottom}} \times h \bigl\}$, is a set of all pixel coordinates inside the bbox of resolution $w \times h$. In our implementation, $\BBB$ is produced by a tuple of the four scalars representing the boundary of the bbox $\bb = (b_{\text{left}}, b_{\text{top}}, b_{\text{right}}, b_{\text{bottom}})$, where $b_{\text{left}}, b_{\text{top}}, b_{\text{right}}, b_{\text{bottom}} \in [0,1]$ specify the bbox relative to the synthesis resolution. The height $h$ and width $w$, are defined by the resolution of the UNet intermediate representation \cite{RombachStablediffusion21}.


\subsection{Spatial Cross Attention Guidance}
\label{ssec:sattn}

The spatial cross attention modules are implemented in the denoising UNet module of \cite{RombachStablediffusion21}. This module finds the cross attention between the query representation $\QQ_s \in \Reals^{N_F \times d_{h} \times d}$ obtained from the SD latent $\zz_t$, and the representations $\KK_s, \VV_s \in \Reals^{N_F \times |W| \times d}$ of the $|W|$ prompt words from the text model, where $d$ is the feature dimension of the keys and queries. Usually $|W| \equiv 77$  when the text embedding model is CLIP \cite{CLIP}. The cross attention map \cite{PromptToPrompt} is then defined as $\bAA_s = \text{Softmax}(\QQ_s\KK_s^{T}/\sqrt{d}) \in \Reals^{N_F \times d_h \times |W|}$,\footnote{Note that this is a ``batch'' matrix multiplication (e.g., the method torch.bmm in PyTorch \cite{paszke2019pytorch}), that is $\CC = \bAA \BB \in \Reals^{b \times m \times n}$, where $\bAA \in \Reals^{b \times m \times p}$, and $\BB \in \Reals^{b \times p \times n}$. Similarly, the transpose operation is $\bAA^\top \in \Reals^{b \times p \times m}.$} where $d_h \equiv w \times h$, defined by the spatial resolution height and width at the specific layer. For simplicity we omit the batch size and the number of 
attention heads \cite{allyouneedattn} in our definition.

As illustrated in the blanched almond-colored section in \FigRef{fig:pipe}, we guide the denoising path by editing the spatial cross attention (e.g., \cite{ma2023directed}) for the attention maps $\bAA_s^{(i)} \in \Reals^{N_F \times d_h}$ associated with a particular prompt word and trailing indices $i \in \III \cup \TTT$. Given $\BBB$, our spatial attention editing is defined by


\begin{equation}
    \text{S}_s(x,y) =
    \left\{\begin{matrix}
        c_s\ g(x, y), \quad (x, y) \in \BBB \\
        0, \quad \text{otherwise},
    \end{matrix}\right.
    \qquad
    \text{W}_s(x,y) =
    \left\{\begin{matrix}
        c_w, \quad (x, y) \in \BBB' \\
        1, \quad \text{otherwise},
    \end{matrix}\right. 
\end{equation}

\noindent where $x, y$ are are the spatial location indices of the attention map and $\BBB'$ is the complement of $\BBB$. $\text{S}_s(\BBB)$ uses a function $g(\cdot,\cdot)$ that ``injects'' attention inside $\BBB$, as illustrated in the gray box in \FigRef{fig:pipe}. The parameters  $c_w \leq 1$, $c_s > 0$ attenuate the attention outside of $\BBB$ and strengthen it inside. We define $g(\cdot,\cdot)$ as a Gaussian window of size $\sigma_x = b_w/2, \sigma_y = b_h/2$, where $b_w = \text{ceil}((b_\text{right} - b_\text{left}) \times w), b_h = \text{ceil}((b_\text{top} - b_\text{bottom}) \times h)$ are the width and the height of $\BBB$. In contrast, $\text{W}_s(\cdot)$ attenuates the attention outside $\BBB$. The bbox $\BBB$ is extended across the entire video sequence through linear interpolation of the keyframes. For example, $\BBB_f = (1 - a) \times \BBB_b + a \times \BBB_e$, where $a = \frac{f}{N_F}$, and $\BBB_b$, $\BBB_e$ denotes the bbox for the beginning and the end of keyframe.

Given the set of indices of subject prompt words $\III$ and trailing maps $\TTT$, each cross-activation component at location $(x,y)$ in $\bAA_s$ is modified as follows:
\begin{equation}
    \bAA_s^{(i)}(x,y) \coloneqq \bAA_s^{(i)}(x,y) \odot \text{W}_s(x,y) + \text{S}_s(x,y),\ \forall i \in \III \cup \TTT,
\label{eq:sa}
\end{equation}
where $\odot$ denotes the Hadamard (element-wise) product that scales the $x,y$ element of the cross-attention map $\bAA_s$ by the corresponding weight in $\text{W}_s(\cdot)$. The result is that the attention in the cross-attention map for the particular prompt word as well as the trailing maps, is stronger in the user-specified bbox region.

\subsection{Temporal Cross-Frame Attention Guidance}
\label{ssec:tattn}

To capture the temporal correlation in the video clip during training, a prevalent approach involves reshaping the latent tensor. This involves shifting the spatial information to the first dimension, a technique employed in VideoLDM \cite{blattmann2023videoldm}. The reshaping is done before passing the hidden activation into the temporal layers, allowing the model to learn about the ``correlation'' of spatial components through the convolutional layers. As shown in blue section in \FigRef{fig:pipe}, the temporal attention map is obtained by $\bAA_m = \text{Softmax}(\QQ_m\KK_m^{T}/\sqrt{d}) \in \Reals^{d_h \times N_F \times N_F}$, where $d_h$ is the spatial dimensions of this tensor,  $\QQ_m \in \Reals^{d_{h} \times N_F \times d}$, and $\KK_m \in \Reals^{d_{h} \times N_F \times d}$.   

What is different from the spatial counterpart is that now $\bAA_m$ learns about the relation between the correlated components across all frames. For instance, $\bAA_m^{(x,y,i,j)}$ denotes the correlation at location $(x,y)$ between frame $i$ and frame $j$. We denote such tensors as $\bAA_m^{(i,j)}(x,y)$ to keep the notation consistent. As seen in our visual investigation (\FigRef{fig:mot}, Right), the background attention is higher when the cross frame attention compares the frames that are temporally far from each other, and the foreground attention is higher when the frames are temporally closer in the video sequence.

To achieve this pattern of activations under user control we design an approach similar to \EqRef{eq:sa} but considering the normalized video temporal distance $d = \frac{|i - j|}{N_F}, i,j \in \{1,...,N_F\}$, the temporal injection function is defined as,
\[
\text{S}_m(x,y) =
    \left\{\begin{matrix}
        (1 - d) \ g(x, y) - d \ g(x, y),\ (x, y) \in \BBB, \\
        0, \quad \text{otherwise}.
    \end{matrix}\right.
\]
Here the normalized video temporal distance $d$ determines the level of the weight injection as a triangular window in time. Values $d \approx 0$ increase the activation inside the bbox. In contrast, when $d \approx 1$, the activation inside the box is \emph{reduced}, approximating the temporal ``anti-correlation'' effect seen in \FigRef{fig:mot}.  The editing by $\text{S}_m(\cdot)$ is performed during the initial $N_M$ steps of the denoising process. 

Then, similarly to \EqRef{eq:sa}, the temporal cross-frame attention map editing is,
\begin{equation}
    \bAA_m^{(i,j)}(x,y) \coloneqq \bAA_m^{(i,j)}(x,y) \odot \text{W}_m(x,y) + \text{S}_m(x,y),
\label{eq:ma}
\end{equation}
where $\text{W}_m(\cdot)$ is defined the same as $\text{W}_s(\cdot)$.

\subsection{Scene compositing}
\label{ssec:comp}

The problem space becomes more complicated for video synthesis with more than one moving subject. Although the parameters $c_s, c_w$ in \EqRef{eq:sa} are specific to a particular subject, they indirectly affect the entire scene through the global denoising. Thus, the choice of these parameters for different subjects might interact and require a parameter search in the number of subjects to find the best synthesis. If the prompt $\PPP$ and bbox $\BBB$ are in conflict then the result might be poor. For instance, a user may specify motion of $\BBB$ from left to right associated with the prompt word ``dog'', while $\PPP$ is given as ``a dog is sitting on the road''. In fact, the dog moves in accordance with the configured bbox, either walking or running.

\begin{figure}[tp]
  \centering
  \includegraphics[width=\linewidth]{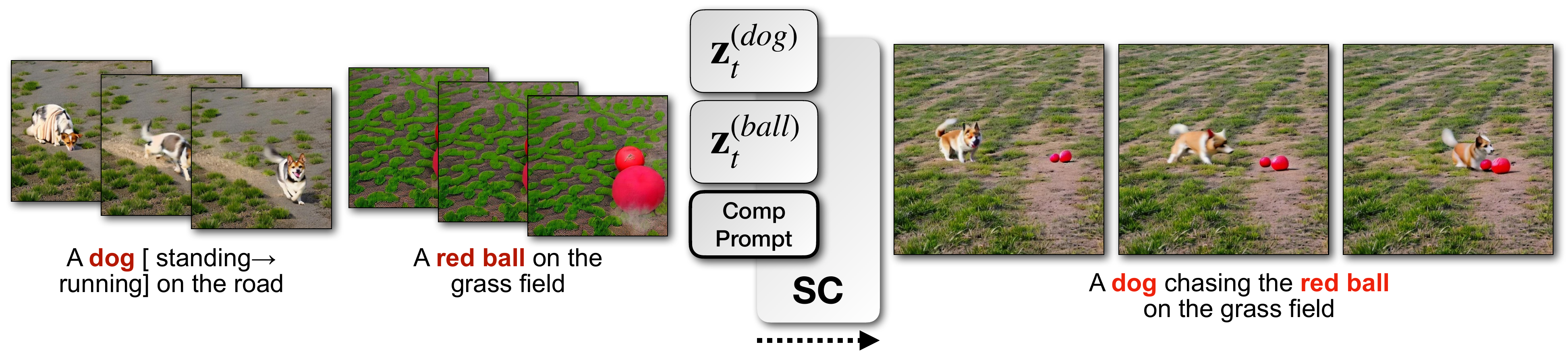}
  \vspace{-1em}
  \caption{\textbf{Scene Compositing.} Given the set of latents generated from our system using a single bbox denoted as $\zz^{(\text{ball})}_{t}$ and $\zz^{(\text{dog})}_{t}$ for the case of prompts related to ball and dog. Then, the scene compositor (SC) produces a synthesis of multiple subjects with the complete prompt and the single subject latents. We refer reader to our supplementary video to view the implemented speed control of the dog.}
  \label{fig:sccomp}
\end{figure}

Considering the reasons above, we follow work such as \cite{ma2023directed, multidiffusion} that combine multiple subjects, each with their own prompt, during the latent denoising. The latents $\zz^{(r)}_t$ for the $r$-th subject are then composited into an overall image latent $\zz_t$ under the control of a ``composed'' prompt, as illustrated in \FigRef{fig:sccomp} and formulated as,

\begin{equation}
    \zz_t(x,y) \coloneqq \frac{1}{R} \sum_{r=0}^{N_R} w \, \zz_t(x,y) + (1{-}w) \, \zz^{(r)}_t\!(x,y),
    \label{eq:compositing}
\end{equation}
where $\forall t \in \{T,...,T{-}N_C\}, \quad (x,y) \in \BBB_r$, where $w \in [0,1]$ determines the weight of linear interpolation between the specific subject latent $\zz^{(r)}_t$ and the composed latent $\zz_t$. 
It is formulated by considering the ratio of the current denoising timestep between $N_C$ and $T$, such that $w = 1 - \big(N_C - (T-t)\big)/N_C$.
At the beginning of the denoising process (so at $t=T$), the compositing fully prioritizes the subject latent $\zz^{(r)}_t$ in each local region in the associated bbox $\BBB_r$. As $t$ decreases, $w$ gradually increases, giving higher priority to composed latent $\zz_t$. This process concludes when $t=T-N_C$, resulting in $w = 1$, which stops using the subject latent in the remaining denoising steps.

\section{Experiments}
\label{sec:exp}

Here we briefly present some experiments and quantitative evaluations of our work. Please see our supplementary materials and the project video for full experiments, including implementation details, limitations, ablations, comparisons, and finer details. The figures show an evenly spaced temporal sampling of frames from the videos. 


\subsection{Main result}

\begin{figure*}
  \centering
  \includegraphics[width=\textwidth]{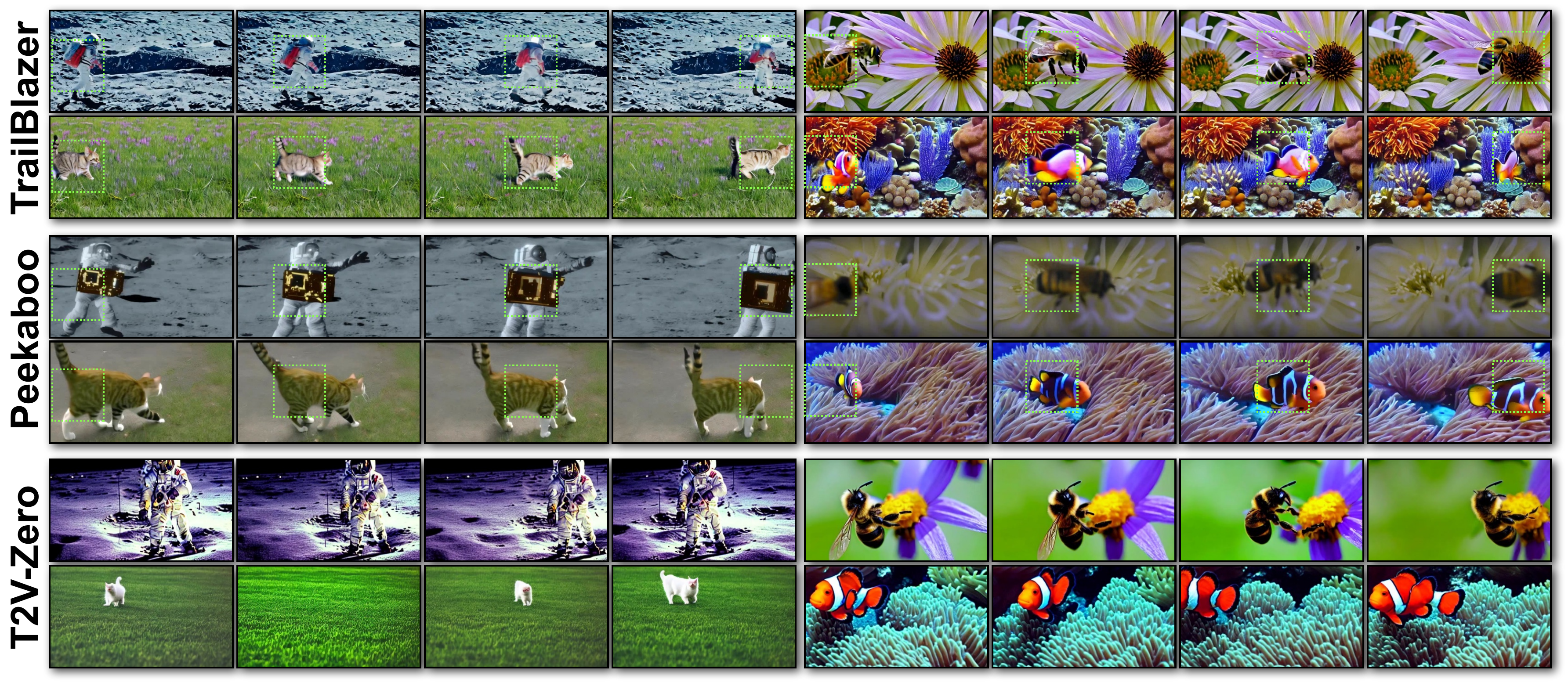}
  \caption{\textbf{Main result: Rigid bbox moving from left to right.} \TitleLong and Peekaboo use identical bboxes, while T2V-Zero uses the corresponding motion vector instead. The same prompt is used across each method. The four prompts used (clockwise from top left): An \textbf{astronaut} walking on the moon; A macro video of a \textbf{bee} pollinating a flower; A \textbf{clown fish} swimming in a coral reef; A \textbf{cat} walking on the grass field. The bold text represents the directed object.}
  \label{fig:main_static}
\end{figure*}

\FigRef{fig:main_static} shows our main result on trajectory control of a  single subject. We compare \textit{TrailBlazer} to T2V-Zero \cite{text2video-zero} and Peekaboo \cite{jain2023peekaboo} using the same prompts, without conditioning guidance (e.g.,~edge or depth maps) to provide a fair comparison. T2V-Zero accepts motion guidance in the form of an (x,y) translation vector. We set this vector to (8,0) to produce horizontal motion. More detailed visual comparisons with Peekaboo under extreme conditions are depicted in \SecRef{supp:sec:peek} of the supplementary materials. The presented results for each method are visually selected as the best of a pool of 10 experiments conducted with different random seeds.

In \FigRef{fig:main_static}, the results are generated from linearly interpolated bboxes starting at the left of the image and moving to the right. The results from TrailBlazer demonstrate anatomically plausible motion of the subject and a more accurate fitting of the subject within the bbox. Further, all subjects (e.g., cat, bee, astronaut, and clown fish) face in the direction that they move. However, this is not a common occurrence in T2V-Zero, as they directly apply the editing operation on the diffusion latents. This approach merely translates the subject without re-orienting it. Although the synthesized subject's motion generally follows the bbox in Peekaboo, it does not fit the bbox well. Occasionally, artifacts may emerge, such as a rectangular object following the astronaut. Moreover, our synthesized background exhibits better visual quality. In the competing methods, the background often appears plain, blurry, or lacks detail behind the subject (e.g., the area behind the subject path in Peekaboo).

\begin{figure*}
  \centering
  \includegraphics[width=\textwidth]{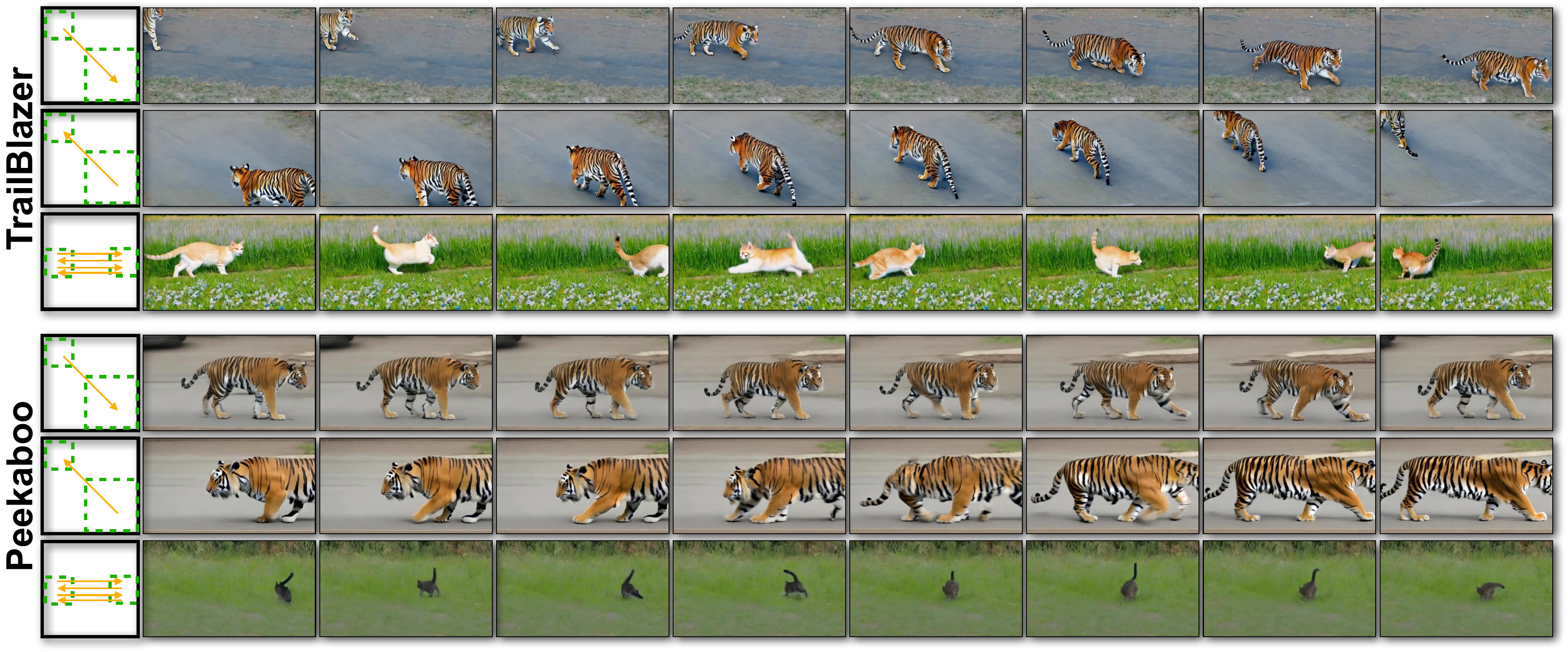}
  \caption{\textbf{Main result: Dynamic moving bbox.} (Top/Middle row): The \textbf{tiger} walking on the street. (Bottom row): The \textbf{cat} running on the grass. The first column illustrates the bbox keyframes in the squared layout, where the green bbox is guided by the almond-colored motion vector. Note that there are five keyframes in the third row of each subfigure, with the bbox located on the left at the initial keyframe. The bbox used in top and the middle row are linearly interpolated with varied sizes. The bbox used in the last row has a static size with 5 keyframes moving back-and-forth. Please refer to the main text for more detail.}
  \label{fig:main_dyn}
\end{figure*}

\FigRef{fig:main_dyn} illustrates the speed control, and dynamically changing the bbox size producing an effect of the subject moving toward or away from the virtual camera. In the top two rows of comparisons, the bbox setup is between the top-left corner and the bottom-right corner. The dynamically changing bbox size is annotated with a green box as illustrated on the left. Note that the generated subjects share the desirable characteristic that the subject naturally faces toward the virtual camera when the bbox transitions from small to large as seen in the top sequence and vice versa in the second sequence from the top.
The results also show a desirable perspective effect. Increasing or reducing the bbox size over time causes the synthesized object to produce the motion of ``coming to'' and ``going away from'' the camera as shown in the tiger example. We believe these effects arise naturally as a result of manipulating a model that was trained on video sequences rather than images. Peekaboo's tiger fails to produce these perspective effects when guided with identical bboxes.


Furthermore, our method adeptly manages fast motion, outperforming Peekaboo in this regard. This is evident in the third row of both subfigures of \FigRef{fig:main_dyn}, which illustrate a cat rapidly running from one side to the other multiple times in the video clip, where $N_f = 24$. \KMtext{More precisely, the bbox is initially positioned at the left (1st keyframe), after which it is moved to the right (2nd keyframe), then left (3rd keyframe), right (4th keyframe), and left (5th keyframe).}

Multi-subject synthesis is generally challenging, particularly when the number of objects exceeds two. We delve into this issue in the supplementary materials. In \FigRef{fig:main_multi}, we present experiments with two subjects, a cat and a dog, guided by the green bbox in the sub-figure. The synthesis of the dog and cat in isolation is depicted in the top row on the left, serving as a sanity check with annotated image frame.
We also show six results combining environment prompts (e.g., ``... on the moon'') after composed prompt (e.g., ``A white cat and a yellow dog running...''). Each experiment demonstrates the flexibility of TrailBlazer in synthesizing subjects under varied environmental conditions. Notably, the interactions between the background and subjects appear plausible, as seen in reflections and splashes in the swimming pool case and consistent shadows across all samples. The results also show some artifacts such as extra limbs that are inherited from the underlying model. 

\begin{figure}[tp]
  \centering
  \includegraphics[width=1.0\linewidth]{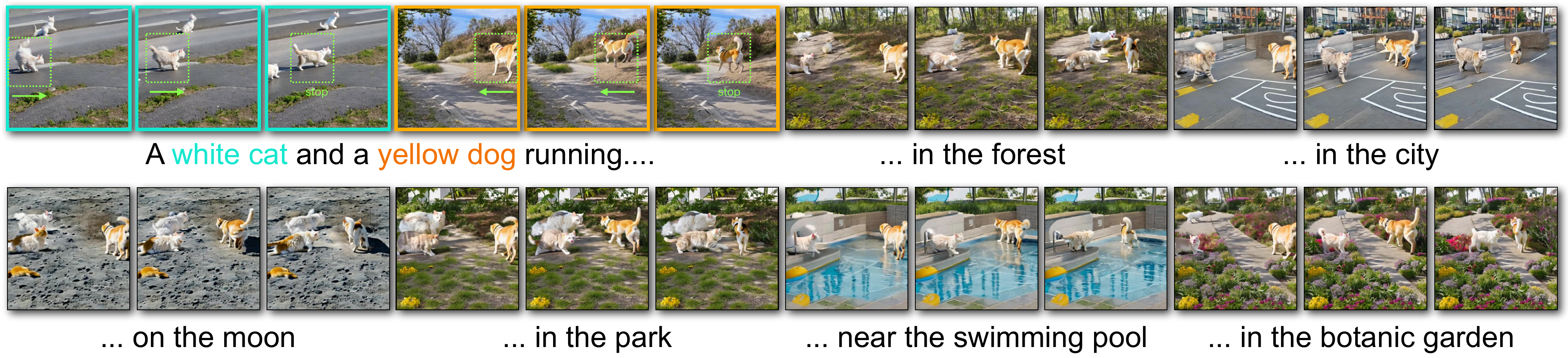}
  \caption{\textbf{Main result: Subjects compositing.} Each set of the three sub-figures representing the first, middle, and the end frame of the synthesized video. The first row on the left with annotated frame shows the video synthesis of the two subjects: ``cat'' and the ``dog'' guided by the bbox directed by the annotated arrows, respectively. Next, each set of results show the varied post-fixed prompt.}
  \label{fig:main_multi}
\end{figure}

\subsection{Quantitative evaluation}
\label{ssec:quant}

Following the methodology in \cite{blattmann2023videoldm, hu2023videocontrolnet, jain2023peekaboo}, we report Fréchet Inception Distance \cite{FID} (FID), Fréchet Video Distance (FVD), Inception Score (IS), Kernel Inception Distance (KID), mean intersection of union (mIoU), and CLIP similarities (CLIPSim) metrics against the random selected 400 videos in AnimalKingdom dataset \cite{Ng_2022_CVPR} on all images of video sequences. As described in the supplementary materials, we evaluate both methods using the prompt set published in \cite{jain2023peekaboo}. The mIoU evaluation utilizes the OWL-ViT-large open-vocabulary object detector \cite{OWLViTlarge} to obtain the bbox of the synthesized subject.

For a fair quantitative evaluation, we generated baseline results using Peekaboo \cite{jain2023peekaboo} and T2V-Zero \cite{text2video-zero} without additional conditioning input. Both TrailBlazer and Peekaboo share the same keyframed bbox, the motion vectors are used for T2V-Zero depending on the tasks below, and we use a 24-frame video sequence as our baseline comparison. We conducted two experiments with the associated random keyframing for our work: \emph{Static bbox}, and \emph{Dynamic bbox}.

The bboxes in the \emph{Static bbox} experiments are constant across all keyframes, where the top left corner is randomly generated in the second quadrant, and the width and height is randomly selected between 25\% to 50\% of the image resolution. This experiments mainly evaluate the method without considering the bbox motion. The result is summarized in Table.~\ref{tab:quant:stt}. As observed, our performance is roughly equivalent across all metrics, while our FVD is significantly lower than that of T2V-Zero and Peekaboo. Motion vector $x=0, y=0$ is used in T2V-Zero.

\begin{table}
  \centering
  \begin{tabular}{c c c c c c c}
    \toprule
    Method & FID($\downarrow$) & FVD($\downarrow$) & IS($\uparrow$) & KID($\downarrow$) & mIoU($\uparrow$) & CLIPSim($\uparrow$)\\
    \midrule
    T2V-Zero \cite{text2video-zero} 
        & 198.45 & 2897.57 & 2.48 $\pm$ 0.31 & 4.39\% & - & \textbf{31.54} \\
    Peekaboo \cite{jain2023peekaboo} 
        & \textbf{159.55} & 1521.27 & 2.27 $\pm$ 0.49 & \textbf{2.75}\% & 0.23 & 31.31 \\
    \TitleLong 
        & 182.28 & \textbf{1220.98} & \textbf{2.91} $\pm$ 0.54 & 3.51\% & \textbf{0.26} & 30.93 \\
    \bottomrule
  \end{tabular}
  \caption{Quantitative results for static bbox.}
  \label{tab:quant:stt}
  \vspace{-2em}
\end{table}

Table.~\ref{tab:quant:dyna} presents the results of the \emph{Dynamic bbox} experiments. The bboxes were generated by randomly specifying 2 to 6 keyframes evenly in the video clip where the location is at the image boundary and its opposite as shown in \FigRef{fig:main_dyn}. The height and width of the particular bbox is between 10\% to 50\%. Thus the location and the size of all bboxes are varied in a video clip. Motion vector $x=8, y=0$ is used in T2V-Zero.

In Table.~\ref{tab:quant:dyna}, the notable improvement is our mIoU score compared to Peekaboo, which can be attributed to the capabilities demonstrated in \FigRef{fig:main_dyn}, showcasing TrailBlazer's proficiency in generating a perspective view with dynamically changing bboxes. In comparison to Peekaboo, TrailBlazer's FID is better, while they exhibit better FVD. This discrepancy may be explained by the nature of the AnimalKingdom \cite{Ng_2022_CVPR} dataset, where creatures typically perform actions in a stationary setting (e.g., birds singing, animals walking). Notably, the running cat motion in \FigRef{fig:main_dyn} is generally absent in their dataset, contributing to the lower FVD score in our case. Our better FID score suggests that the individual frame quality in our video clip is better.

\begin{table}
  \vspace{2em}
  \centering
  \begin{tabular}{c c c c c c c}
    \toprule
    Method & FID($\downarrow$) & FVD($\downarrow$) & IS($\uparrow$) & KID($\downarrow$) & mIoU($\uparrow$) & CLIPSim($\uparrow$)\\
    \midrule
    T2V-Zero \cite{text2video-zero} 
        & 153.36 & 2350.69 & \textbf{4.51} $\pm$ 1.36 & 3.01\% & - & 31.32 \\
    Peekaboo \cite{jain2023peekaboo} 
        & 165.88 & \textbf{1571.04} & 2.16 $\pm$ 0.57 & \textbf{2.86}\% & 0.25 & \textbf{31.92}\\
    \TitleLong 
        & \textbf{148.18} & 1721.08 & 2.45 $\pm$ 0.61 & 3.12\% & \textbf{0.37} & 30.71 \\
    \bottomrule
  \end{tabular}
  \caption{Quantitative results for dynamic bbox.}
  \label{tab:quant:dyna}
  \vspace{-2em}
\end{table}

In summary, the objective scores in Tables 1, 2 do not give a clear ordering of methods. However, recall that our goal is \emph{controlling movement}. TrailBlazer achieves this, showing significantly better mIoU scores. Equally important, TrailBlazer shows improved subjective movement, with moving objects facing in plausible directions and having realistic motion. Please refer to our video.




\section{Conclusion}
\label{sec:cond}

We have addressed the problem of controlling the motion of objects in a diffusion-based text-to-video model. Specifically, we introduced a combined spatial and temporal attention guidance algorithm, \textit{TrailBlazer}, operating in the pre-trained ZeroScope model. 
The spatial location of a subject can be guided through simple bounding boxes. 
Bounding boxes and prompts can be animated via keyframes, enabling users to alter the trajectory and coarse behavior of the subject along the timeline. The resulting subject(s) fit seamlessly in the specified environment, providing a viable approach to video storytelling by casual users.  Our approach requires no model finetuning, training, or online optimization, ensuring computational efficiency and a good user experience. Lastly, the results are natural, with desirable effects such as perspective, motion with the correct object orientation, and the interactions between object and environment arising automatically.

\clearpage
\setcounter{page}{1}
\onecolumn

\begin{center}
  \Large \textbf{\thetitle}\\
  \large \vspace{0.5em} Supplementary Material
  \vspace{1.0em}
\end{center}



\section{Implementation}
\label{sec:impl}

In this section we describe details of  implementation in \textit{TrailBlazer}, including the core library, hyperparameters, and other pertinent information. Our method is developed using PyTorch 2.01 \cite{paszke2019pytorch}, and the Diffusers library version 0.21.4 from Huggingface \cite{huggingface}. We override the Diffusers pipeline \texttt{TextToVideoSDPipeline} to produce our implementation.

Parameters are selected as follows: We use classifier-free guidance with a strength of 9, conduct 40 denoising steps, and maintain a video resolution of 512x512 for the conventional stable diffusion backward denoising process. For all the comparisons with Peekaboo \cite{jain2023peekaboo} we use their official repository\footnote{\href{https://github.com/microsoft/Peekaboo}{https://github.com/microsoft/Peekaboo}} at the commit 6564274 (12 Feb 2024). In our comparisions we utilize a resolution of 576x320 as employed in the Peekaboo code to ensure fair assessment.

Regarding the parameters specific to our proposed method, the majority of our results are generated using the default values outlined as follows: We execute 5 editing steps for both spatial and temporal attention, denoted as $N_S \equiv N_M \equiv 5$. The editing coefficients $c_w \equiv 0.001$ and $c_s \equiv 0.1$ are used in both spatial and temporal attention in most cases. The number of trailing attention maps $|\TTT|$ is the only parameter that needs to be tuned. Generally, $10 \leq |\TTT| \leq 20$ yields satisfactory results in practice and we set $|\TTT| \equiv 15$ for our paper results.

As highlighted in \SecRef{sec:intro}. in the main text, we adapt the pre-trained ZeroScope\footnote{\cite{huggingface}:cerspense/zeroscope\_v2\_576w} \cite{zeroscope} T2V model. This model is fine-tuned from the initial weights of ModelScope \cite{luo2023videofusion}\footnote{\cite{huggingface}:damo-vilab/modelscope-damo-text-to-video-synthesis} utilizing nearly ten thousand clips, each comprising 24 frames as training data. Consequently, we adhere to the recommended practice of setting the length of the synthesized sequence to 24 frames, drawing insights from user experiences shared in relevant blogs. \footnote{\href{https://zeroscope.replicate.dev/}{https://zeroscope.replicate.dev/}}

Spatial attention editing is performed at several resolutions with a module with the following architecture:
\begin{footnotesize}
\begin{small}\begin{verbatim}
    transformer_in.transformer_blocks.0.attn2
    down_blocks.0.attentions.0.transformer_blocks.0.attn2
    down_blocks.0.attentions.1.transformer_blocks.0.attn2
    down_blocks.1.attentions.0.transformer_blocks.0.attn2
    down_blocks.1.attentions.1.transformer_blocks.0.attn2
    down_blocks.2.attentions.0.transformer_blocks.0.attn2
    down_blocks.2.attentions.1.transformer_blocks.0.attn2
    up_blocks.1.attentions.0.transformer_blocks.0.attn2
    up_blocks.1.attentions.1.transformer_blocks.0.attn2
    up_blocks.1.attentions.2.transformer_blocks.0.attn2
    up_blocks.2.attentions.0.transformer_blocks.0.attn2
    up_blocks.2.attentions.1.transformer_blocks.0.attn2
    up_blocks.2.attentions.2.transformer_blocks.0.attn2
    up_blocks.3.attentions.0.transformer_blocks.0.attn2
    up_blocks.3.attentions.1.transformer_blocks.0.attn2
    up_blocks.3.attentions.2.transformer_blocks.0.attn2
\end{verbatim}\end{small}
\end{footnotesize}

For temporal attention editing, we found that a multiple-resolution approach was not necessary and produced unpredictable results. Instead, temporal attention editing uses a single layer:
\begin{footnotesize}
\begin{verbatim}
    mid_block.attentions.0.transformer_blocks.0.attn2
\end{verbatim}
\end{footnotesize}

Following \cite{jain2023peekaboo}, the following prompt set is used for the experiments in our quantitative comparison in \SecRef{ssec:quant} in the main text. We include it here only for completeness. The prompt word(s) in bold case is the subject for positioning: 

\begin{itemize}
    \item A \textbf{woodpecker} climbing up a tree trunk.
    \item A \textbf{squirrel} descending a tree after gathering nuts.
    \item A \textbf{bird} diving towards the water to catch fish.
    \item A \textbf{frog} leaping up to catch a fly.
    \item A \textbf{parrot} flying upwards towards the treetops.
    \item A \textbf{squirrel} jumping from one tree to another.
    \item A \textbf{rabbit} burrowing downwards into its warren.
    \item A \textbf{satellite} orbiting Earth in outer space.
    \item A \textbf{skateboarder} performing tricks at a skate park.
    \item A \textbf{leaf} falling gently from a tree.
    \item A \textbf{paper plane} gliding in the air.
    \item A \textbf{bear} climbing down a tree after spotting a threat.
    \item A \textbf{duck} diving underwater in search of food.
    \item A \textbf{kangaroo} hopping down a gentle slope.
    \item An \textbf{owl} swooping down on its prey during the night.
    \item A \textbf{hot air balloon} drifting across a clear sky.
    \item A \textbf{red double-decker bus} moving through London streets.
    \item A \textbf{jet plane} flying high in the sky.
    \item A \textbf{helicopter} hovering above a cityscape.
    \item A \textbf{roller coaster} looping in an amusement park.
    \item A \textbf{streetcar} trundling down tracks in a historic district.
    \item A \textbf{rocket} launching into space from a launchpad.
    \item A \textbf{deer} standing in a snowy field.
    \item A \textbf{horse} grazing in a meadow.
    \item A \textbf{fox} sitting in a forest clearing.
    \item A \textbf{swan} floating gracefully on a lake.
    \item A \textbf{panda} munching bamboo in a bamboo forest.
    \item A \textbf{penguin} standing on an iceberg.
    \item A \textbf{lion} lying in the savanna grass.
    \item An \textbf{owl} perched silently in a tree at night.
    \item A \textbf{dolphin} just breaking the ocean surface.
    \item A \textbf{camel} resting in a desert landscape.
    \item A \textbf{kangaroo} standing in the Australian outback.
    \item A colorful \textbf{hot air balloon} tethered to the ground.
\end{itemize}




\section{Ablations}
\label{sec:ablation}

We conduct ablation experiments on the number of trailing attention maps and the number of temporal steps.

\begin{figure}[tp]
  \centering
  \includegraphics[width=\linewidth]{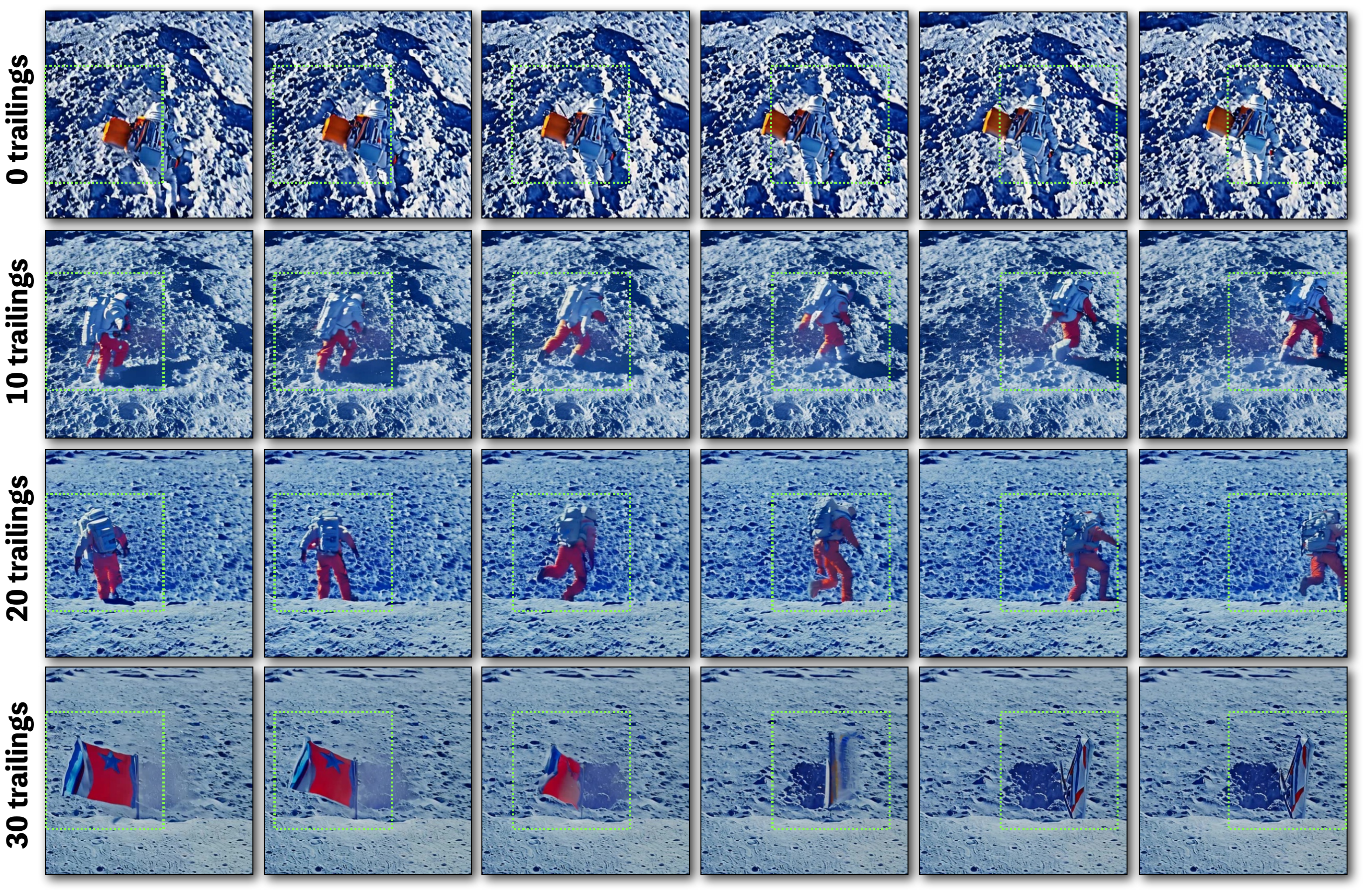}
  \caption{\textbf{Ablation: Trailing maps.}
  The rows from top to bottom show the video synthesis with 0 (no trailing maps), 10, 20, and 30 trailing maps. Prompt: ``The \textbf{astronaut} walking on the moon'', where ``astronaut'' is the directed subject. The number of temporal edit steps is five  in all cases.}
  \label{fig:abl_trailings}
  \vspace{-2em}
\end{figure}

\textbf{Trailing attention maps.} \FigRef{fig:abl_trailings} shows an ablation varying the number of trailing attention maps used in our spatial cross attention process, where the top row shows our method without trailing attention maps (e.g, $|\TTT|=0$) to the bottom row (e.g., $|\TTT|=30$). The guided bbox is annotated with green bbox moving from left to right. It is observed that the astronaut remains static at the image center without the trailing attention maps. In contrast, the synthesis with a large number of trailing attentions can lead to failed results such as a flag rather than the intended astronaut. A good number of edited trailing attention maps is between $|\TTT|=10$ and $|\TTT|=20$.

\begin{figure}[tp]
  \centering
  \includegraphics[width=\linewidth]{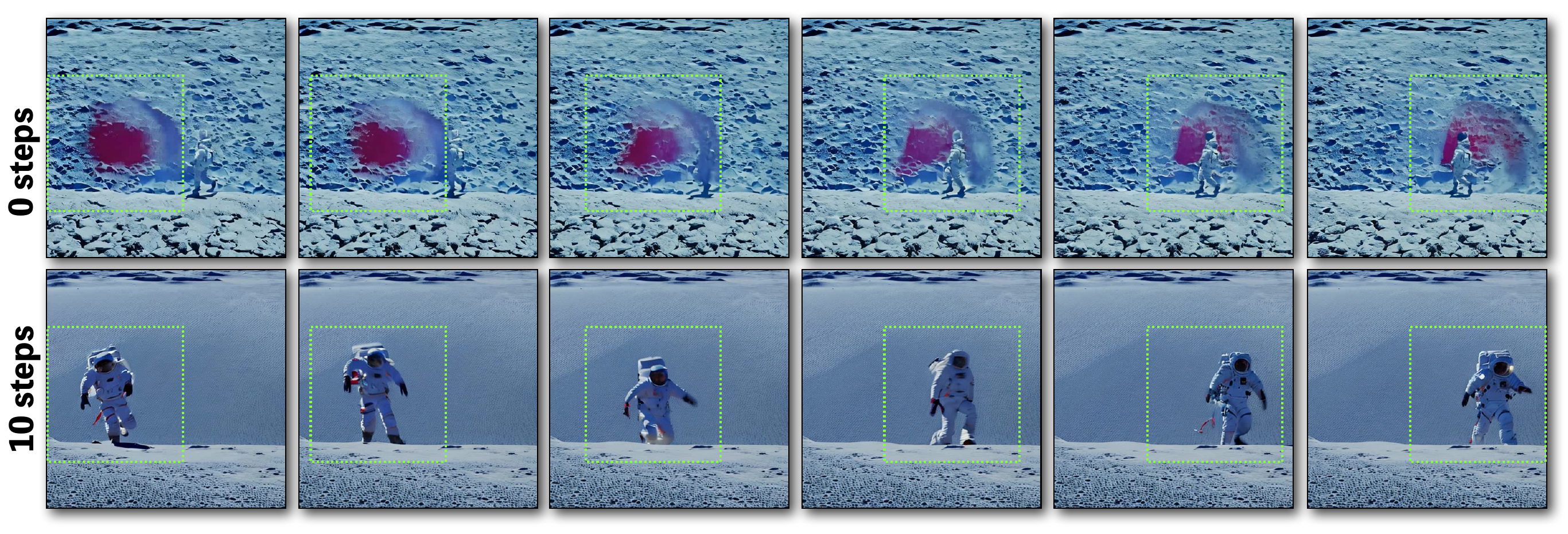}
  \vspace{-0.3em}
  \caption{\textbf{Ablation: Temporal edits.} Following up the experiments in \FigRef{fig:abl_trailings}, the ablation test on the temporal attention editing with varied number of steps of the first and last frame of video reconstruction, shown at the left/right of each set of experiments. (Left/Right): No temporal attention editing, and 10 steps editing, respectively. The number of trailing is 10 for the two cases.}
  \label{fig:abl_temp}
\end{figure}

\textbf{Temporal attention editing.} We further show an ablation test in \FigRef{fig:abl_temp} with varied number of temporal attention editing steps. We take the case of the astronaut from \FigRef{fig:abl_trailings} with $|\TTT|=10$, and set $N_M = 0$ (no editing steps), and $N_M = 10$. The result with $N_M = 0$ shows a red blob moving from left to right. The value $N_M = 10$ gives satisfactory result on the astronaut, but the background along the bbox path is missing. From these results we see that a reasonable balance between spatial and the temporal attention editing must be maintained, while extreme values of either produce poor results. An intermediate value such as $N_M=5$ used in most of our experiments produces the desired result of an astronaut moving over a moon background.

\section{Limitations}
\label{ssec:limits}
\begin{figure}[tp]
  \centering
  \includegraphics[width=\linewidth]{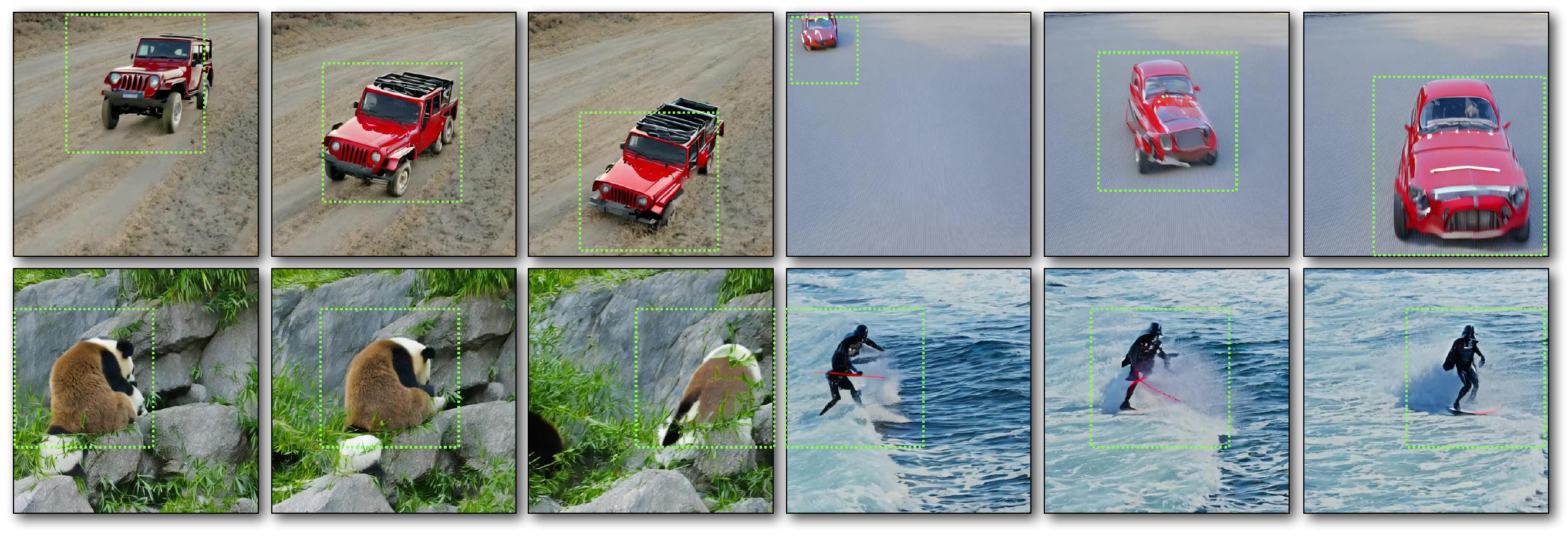}
  \vspace{-0.5em}
  \caption{\textbf{Failure cases.} 
  Prompts used in subfigures: ``\textbf{A red jeep} driving on the road'', ``\textbf{A red car} driving on the highway'', ``\textbf{a panda} eating bamboo'', and ``\textbf{Darth Vader} surfing in waves'', where the bold prompt word is the directed subject.}
  \label{fig:limit}
\vspace{-1em}
\end{figure}

Our method shares and inherits  common failure cases of the underlying diffusion model.
Notably, at the time of writing, models based on CLIP and Stable Diffusion sometimes generate deformed objects and struggle to generate multiple objects and correctly assign attributes (e.g.~color) to objects. We show some failures in \FigRef{fig:limit}. For instance, we requested a red jeep driving on the road but the synthesis shows it sinking into a mud road. The panda example shows the camera moving instead of the panda itself. The red car has implausible deformation, and Darth Vader's light saber turns into a surf board. The length of the resulting video clips is restricted to that produced by the pre-trained model, for instance, the 24 images in the case of ZeroScope. This is not a crucial limitation, as movies are commonly (with some exceptions!) composed of short ``shots'' of several seconds each.  The bbox guides object placement without precisely constraining it. This is an advantage as well, however, since otherwise the user would have to specify the correct x-y aspect ratio for objects, a complicated task for non-artists.

\newpage
\section{Further comparison with Peekaboo}
\label{supp:sec:peek}

\begin{figure*}
  \centering
  \includegraphics[width=\textwidth]{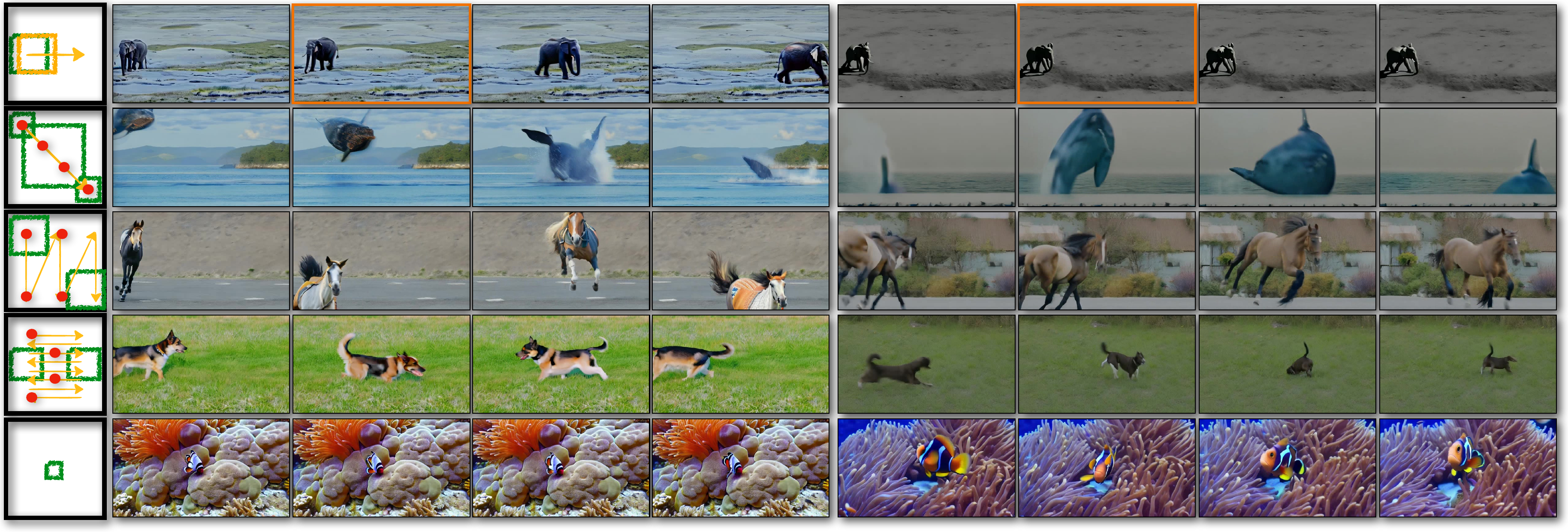}
  \caption{\textbf{Extreme comparison: Various conditions.} \TitleLong (left) and Peekaboo (right) use identical bboxes, while T2V-Zero uses the corresponding motion vector instead. The same prompt is used across each method. The five prompts used from top: An \textbf{elephant} walking on the moon; a photorealistic \textbf{whale} jumping out of water while smoking a cigar; A \textbf{horse} galloping fast on a street; A \textbf{dog} is running on the grass;  A \textbf{clownfish} swimming in a coral reef. The first column at left displays the bbox, and its trajectory. For the sequences with complex motion (2nd, 3rd, 4th row), the frames shown in the figure are denoted by the red dots along the trajectory in the first column. The orange bbox in the first row represents the starting motion of the elephant running to the right near the end of video clip. For additional details, please refer to the accompanying text.}
  \label{fig:extreme}
\end{figure*}

This section provides a comprehensive comparison between \textit{TrailBlazer} and Peekaboo. In \FigRef{fig:extreme}, we explore additional experiments encompassing various scenarios related to bbox size, location, and their combinations under extreme conditions. From top row to bottom: 1) Extremely fast motion by the timing of the second keyframe; 2) Rapid size changes along the bbox trajectory; 3) Zigzag trajectory; 4) Extremely fast motion through numerous keyframes;  and 5) Extremely small bbox. \textbf{Please refer to our the supplementary video to examine the motion for each of the following figures.} 

As depicted in \FigRef{fig:extreme}, TrailBlazer excels in most extreme scenarios for the synthesized subject's location, motion speed, and identity. For example, our representation of an elephant maintains a stationary position for the initial 75\% of the video before initiating movement then runs to the right. The whale gracefully descends into the ocean during the latter part of its jumping motion. The horse accurately follows a zigzag path, simulating a galloping motion. Remarkably, the dog seamlessly follows a large number of keyframes (8 keyframes) within a 24-frame video clip, covering the distance from one boundary to the opposite in approximately 2 frames. The clownfish fits into a tiny bbox. These successes are generally not evident in the Peekaboo \cite{jain2023peekaboo} method.

The metrics presented in Table.~\ref{tab:extreme} are derived from the analysis of experiments shown in \FigRef{fig:extreme} using the AnimalKingdom dataset \cite{Ng_2022_CVPR}, as described in our main text. The mIoU and FID in TrailBlazer surpasses Peekaboo, indicating that our method excels at effectively generating the subject in extreme conditions. Notably, as shown in Table.~\ref{tab:extreme}, our mIoU is approximately twice the value of Peekaboo. As mentioned in the main text, we believe the FVD in TrailBlazer is worse than Peekaboo in this section because the AnimalKingdom dataset does not contain the varied and extreme motion that we used in our experiments.

\begin{table}
  \centering
  \begin{tabular}{c c c c c c c}
    \toprule
    Method & FID($\downarrow$) & FVD($\downarrow$) & IS($\uparrow$) & KID($\downarrow$) & mIoU($\uparrow$) & CLIPSim($\uparrow$)\\
    \midrule
    Peekaboo \cite{jain2023peekaboo} 
        & 258.34 & \textbf{2863.77} & 1.25 $\pm$ 0.24 & 6.71\% & 0.19 & \textbf{30.89} \\
    \TitleLong 
        & \textbf{215.37} & 3416.41 & \textbf{1.47} $\pm$ 0.58 & \textbf{5.08}\% & \textbf{0.39} & 29.41 \\
    \bottomrule
  \end{tabular}
  \caption{Quantitative results for static bbox.}
  \label{tab:extreme}
  \vspace{-2em}
\end{table}

\newpage
\section{Subject Morphing}
\label{sec:morphing}

Subject morphing involves blending semantics for generating images and videos. To our best knowledge, \textit{TrailBlazer} is  first in demonstrating subject morphing by prompt manipulation in the video diffusion domain. Related concepts have earlier been shown for image generation in MagicMix \cite{Magicmix} with, for example, the "corgi coffee machine".

While subject morphing through prompt embedding interpolation may seem less intuitive for real-world applications, it is widely used in the entertainment industry for example for superheroes (e.g., the She-Hulk can transform from a human to a monstrous character). For general usage, it could potentially serve as an entry point for generating new content that is more efficient than using a single prompt, particularly due to the limitations of CLIP \cite{CLIP}. For example, it might be challenging to generate a ``fish-like'' cat using the prompt "A fish-like cat walking on the grass" with a diffusion model. Instead, it would be easier to accomplish this goal by combining the prompt embeddings from ``A fish swimming on the grass'' and ``A cat walking on the grass.''

\begin{figure*}[!htb]
  \centering
  \includegraphics[width=1.0\textwidth]{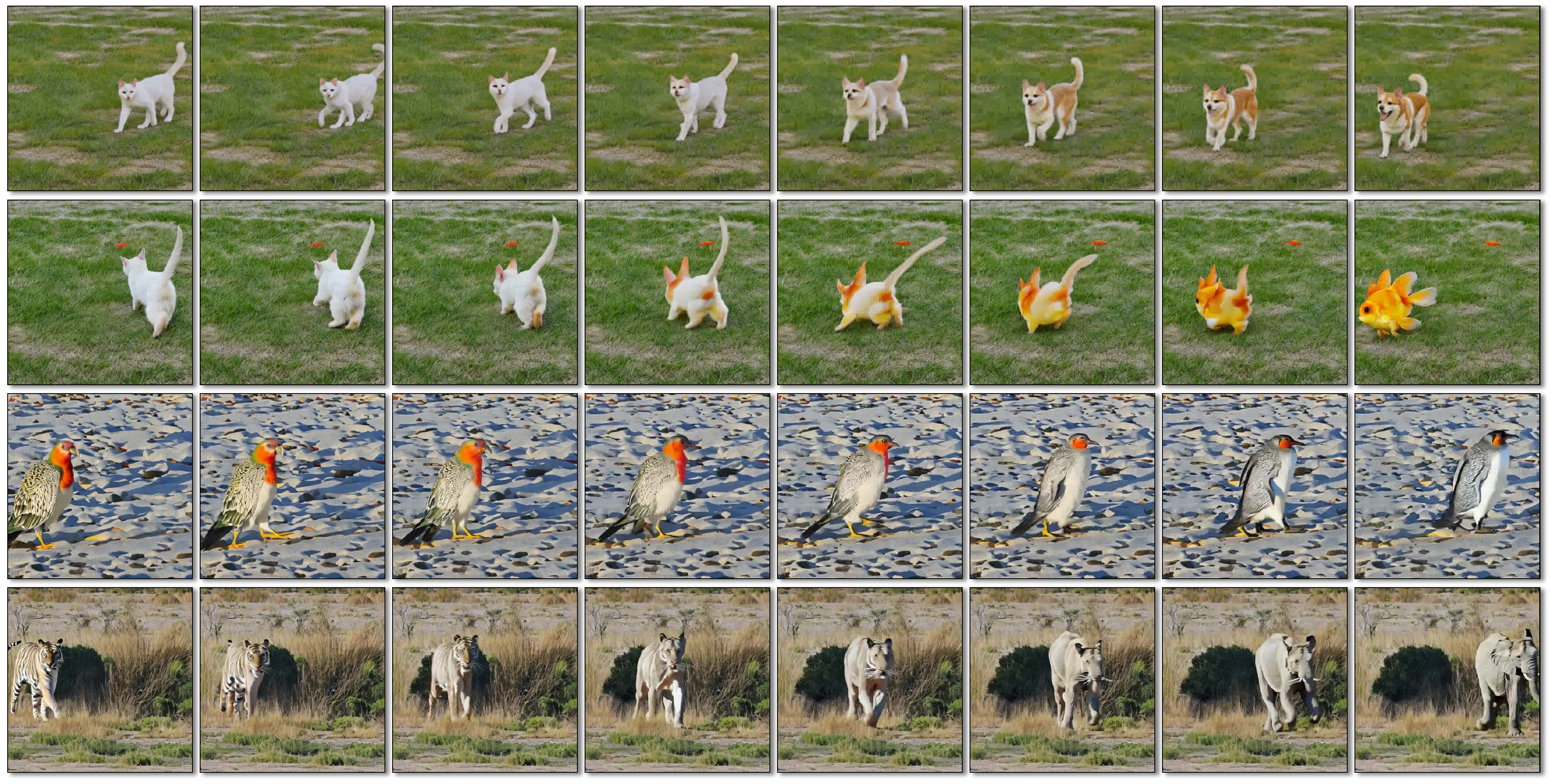}
  \caption{\textbf{Subject Morphing.} The prompts used starting from the first row: ``A [cat $\rightarrow$ dog] walking on the grass'', ``A [cat walking $\rightarrow$ fish swimming] on the grass'', ``A [parrot $\rightarrow$ king penguin] walking on the beach'', and ``A [tiger $\rightarrow$ elephant] walking in the wild park.''. Please refer to the text for more detail.}
  \label{fig:morphing}
\end{figure*}

\FigRef{fig:morphing} illustrates the morphing outcomes generated by \textit{TrailBlazer}. All results are generated using default hyperparameter settings and involve linear interpolation of the prompt embeddings across video frames. The animated bounding boxes shift from right to left in the top two rows, and from left to right in the bottom two rows.

The outcome depicted in \FigRef{fig:morphing} demonstrates that morphing in a video clip can transition smoothly from one identity to another without significant artifacts. Notably, it avoids unrealistic deformations such as generating new joints in unexpected body parts (e.g., a tail on the head) or transforming one animal feature into another (e.g., an eye into an ear). Additionally, the subjects follow exactly the same motion in the synthesis (e.g., walking) across video frames.

\newpage
\section{Comprehensive ablations}
\label{sec:abla}

Given the limited space in the primary text, here we offer more supplementary ablation tests to substantiate our proposed approach. Broadly, we illustrate the impact of the spatial and temporal placement of guidance bounding boxes \emph{(bboxes)} on the overall result quality, exploring the effect of various bbox speed and size choices directed by user keyframing. To see details, please zoom in to the experiment images, and \textbf{especially refer to our supplementary video.}

\begin{figure*}[!htb]
  \centering
  \includegraphics[width=1.0\textwidth]{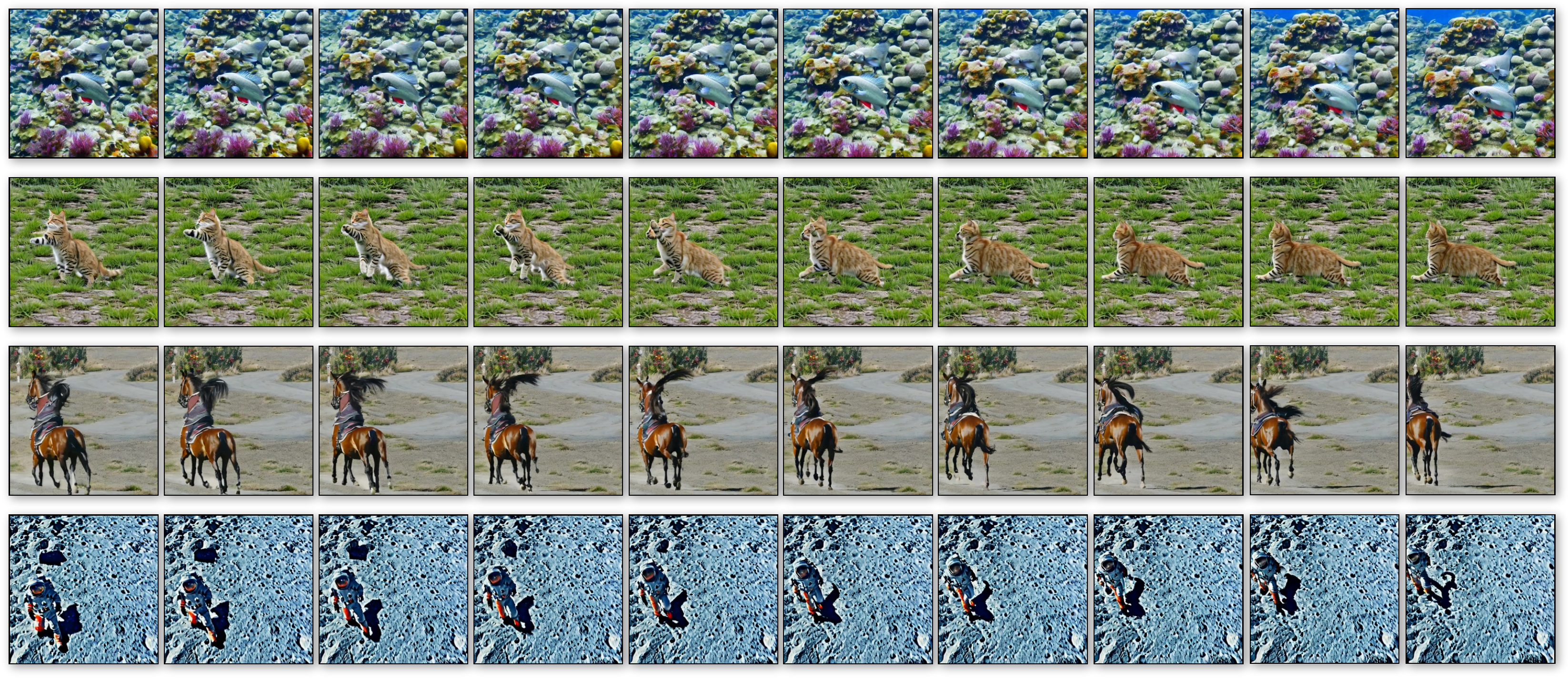}
  \caption{\textbf{Baseline results.} Each row shows equally-spaced frames sampled from a video generated using ZeroScope \emph{without applying our trajectory control approach}. The prompts used starting from the first row: ``A fish swimming in the sea'', ``The cat running on the grass field'', ``The horse galloping on the road'', and ``An astronaut walking on the moon''. These prompts are reused in subsequent examples in these supplementary results.}
  \label{fig:nodd}
\end{figure*}

\FigRef{fig:nodd} illustrates video synthesis using the pre-trained ZeroScope model
\textbf{without} applying our approach. Broadly, all the synthesized results exhibit fine details with plausible temporal coherence as would be seen in a real video featuring relatively slow motion. However, several side effects may be introduced alongside this realism. For example, the synthesized subject is often positioned in the same general area near the center of the images regardless of portrayed motion, and subjects like a galloping horse do not conveying the notion of speed. Additionally, artifacts such as extra or missing limbs (e.g., the cat in the second row) or other implausible results occasionally occur.

\newpage
\subsection{Exploration and Ablation: Varied static bbox sizes}

\begin{figure*}[!htb]
  \centering
  \includegraphics[width=1.0\textwidth]{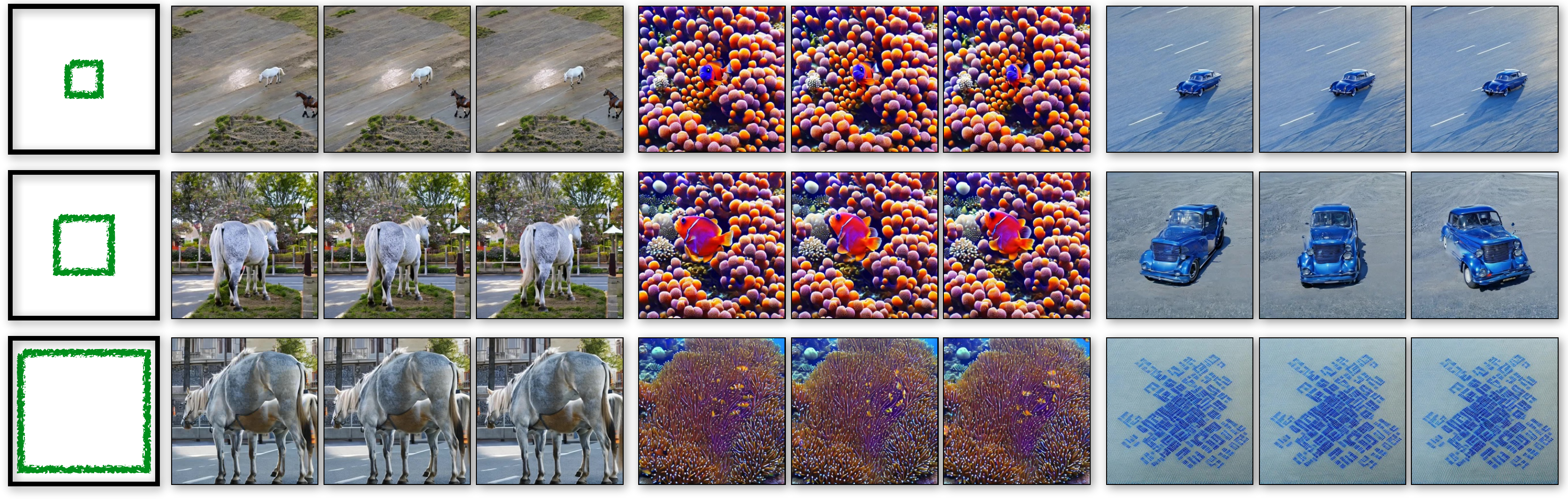}
  \caption{\textbf{Static bbox sizes.} Each row shows the result of a static square bbox positioned at the center, where the width and height are 25\%, 50\%, and 90\% of the original image size (represented by the the green square on the left). The prompts used in the three sets of the experiments are: ``The \textbf{white horse} standing on the street'', ``The \textbf{fish} swimming in the sea'', and ``The \textbf{blue car} running on the road''.}
  \label{fig:supp_static_bbox}
\end{figure*}

\FigRef{fig:supp_static_bbox} shows the effect of the size of the bbox without considering motion.  The results indicate that the bbox size significantly influences the outcome. In extreme cases, the top row illustrates that a smaller bbox may yield unexpected entities in the area (e.g., white smoke next to the horse) or information leakage to the neighboring area (e.g., the blue attribute affecting the road). In contrast, the bottom row demonstrates that a overly large bbox can lead to broken results in general (e.g., the fish disappearing into the coral reef, and the strange blue pattern in place of the expected blue car). We expect this may be in large part due to the centered-object bias \cite{szabo-centeredobjects-21} in the pre-trained model's training data.

Our recommended bbox size falls within the range of 30\% to 60\% for optimal reconstruction quality. Note that very small- or large-sized bboxes can still be employed in our approach, but they are best specified for a particular frame rather than the entire sequence. This is demonstrated, for example, in \FigRef{fig:supp_dyn_bbox} guiding the swimming fish.

\newpage
\subsection{Exploration and Ablation: Varied dynamic bbox sizes}

\begin{figure*}[!htb]
  \centering
  \includegraphics[width=1.0\textwidth]{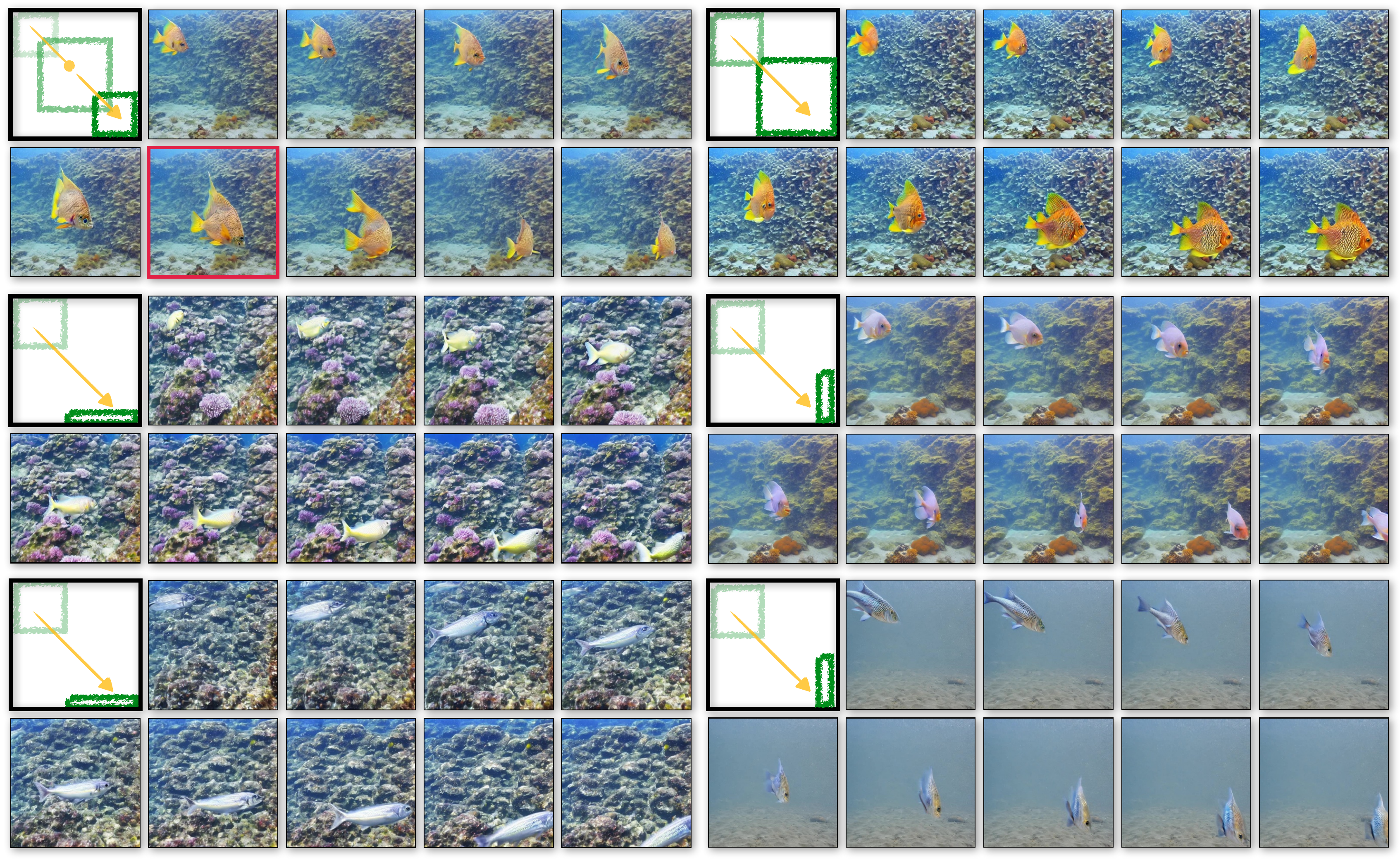}
  \caption{\textbf{Dynamic bbox sizes.} The result showcases six synthesized video sequences with the subject directed by the yellow arrow starting at the position indicated by green bbox. The number of the bboxes corresponding to the number of keyframes used in the experiment is, clockwise from top-left, $|\KKK| =$ 3, 2, 2, 2, 2, and 2, respectively. The prompt used in each result: ``The [X] swimming in the sea'', where ``[X]'' denotes the ``fish'' for the first and second rows, and ``sardine'' for the third row.}
  \label{fig:supp_dyn_bbox}
\end{figure*}

\FigRef{fig:supp_dyn_bbox} demonstrates video synthesis with a dynamically changing bbox size. In the top-left example, the bbox grows larger and then shrinks, resulting in a perspective effect where the fish swims towards the camera and then away from it. The frame highlighted in red indicates the middle keyframe with a large bbox. This aligns with our main text results in \FigRef{fig:main_dyn}, showcasing that the animated tiger and car respect the bbox size. The top-right example is a comparison to the top-left, portraying the fish only swimming toward the camera.

The second and the third rows show a comparison of the same bbox condition with the prompt words ``fish'' (second row), and ``sardine'' (third row), respectively. 
This experiment aims to assess how well our method adapts to large bbox size variations, represented by the short/wide target bbox on the left and tall/thin target bbox on the right. The result on the left indicates that the output from the ``fish'' prompt does not adequately conform to the short-wide aspect ratio of the bounding box, whereas the result from the ``sardine'' prompt can more closely adjust to the desired bbox thanks to the elongated shape of the sardine.
Conversely, in the experiment on the right, both ``fish'' and ``sardine'' perform well with the tall/thin bounding box, since the tall aspect ratio can be satisfied by a fish facing directly toward or away from the camera. In general we expect that the obtained results will mimic the situations found in ZeroScope's training data,
while views that are outside the typical data (such as a fish swimming vertically, or a horse at the top of the image) will be difficult to synthesize.

As with all our results, we see that the guided subject \emph{approximately} follows the specified bounding box, but does not exactly lie within the bbox.
While this is a disadvantage for some purposes, we argue that it is also an advantage for casual users -- if the subject exactly fit the bounding box it would require the user to imagine the correct aspect ratio of the subject under perspective (a difficult task for a non-artists) as well as do per-frame animation of the bbox to produce the oscillating motion of the swimming fish seen here.

\newpage
\subsection{Exploration and Ablation: Speed control with multiple keys}

\begin{figure*}[!htb]
  \centering
  \includegraphics[width=1.0\textwidth]{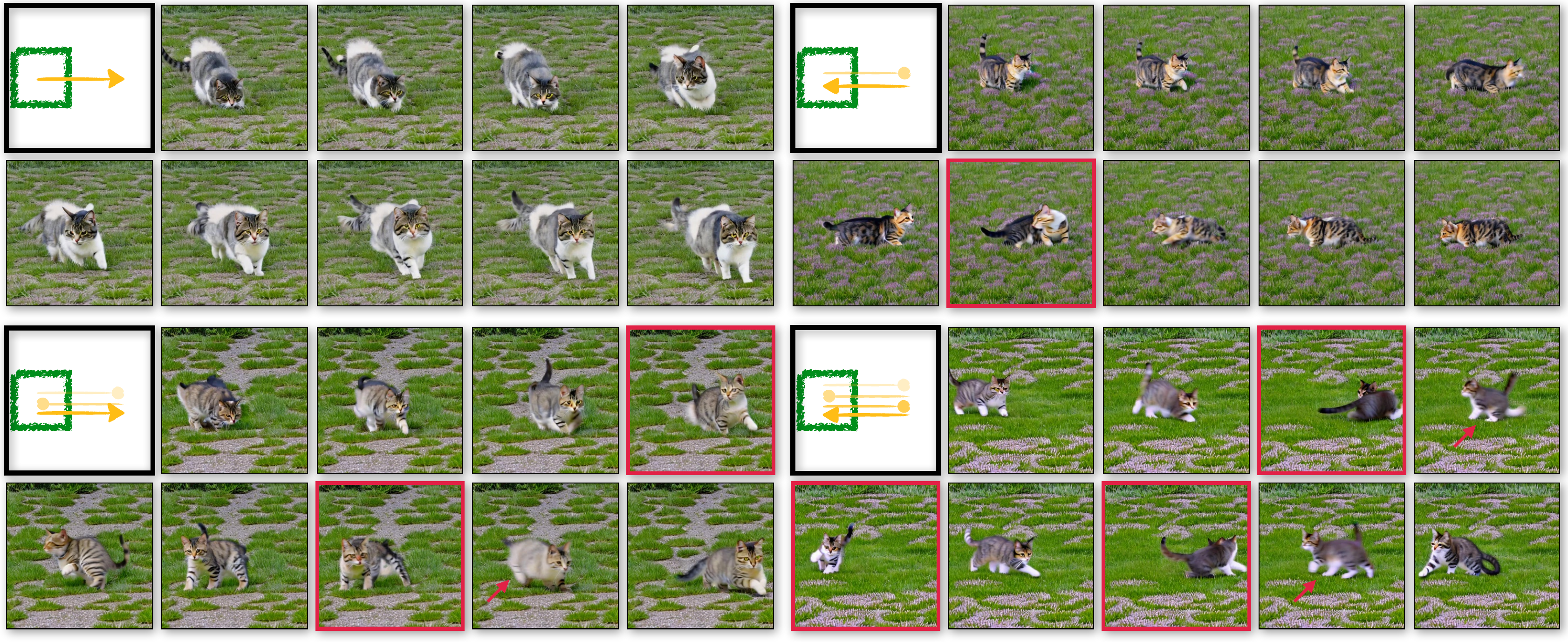}
  \caption{\textbf{Speed Test: number of keyframes.} This result shows four synthesized video sequences with the cat's motion directed according to the yellow arrows starting from the position indicated by green bbox. The number of the arrows denotes the number of keyframes (excluding the start/end keyframes) used in each experiment. Specifically, starting from the top-left and proceeding in left/right top/down (English reading) order, there are
  $|\KKK| = $ 2, 3, 4, and 5, keyframes, respectively. The frames highlighted with red correspond to the user-specified keyframes,  excluding the start and end keyframes. The prompt used for all experiments is ``A \textbf{cat} running on the grass field''. The red arrows in the bottom-right example shows the introduced motion blur representing fast-moving speed.}
  \label{fig:supp_mul_keys}
\end{figure*}

\FigRef{fig:supp_mul_keys} demonstrates controlling the subject's speed through varying the number of keyframes in the video synthesis. Given the recommended sequence length $N_f = 24$ for ZeroScope, we show the result of adding different
keyframes in between the start and end keyframes at the left/right image boundary, simulating the cat running back and forth on the grass field. It is clear that the cat moves relatively naturally according to the motion flow indicated by the yellow arrows. For instance, the cat looks back first before turning around, rather than showing an unnatural motion where the position of the head and tail is instantaneously swapped. As the cat moves faster, motion blur also introduced in the result annotated with red arrows. We found that this motion blur is hard to eliminate using negative prompts.

\newpage
\subsection{Exploration and Ablation: Controlling speed with different placement of a single keyframe}

\begin{figure*}[!htb]
  \centering
  \includegraphics[width=1.0\textwidth]{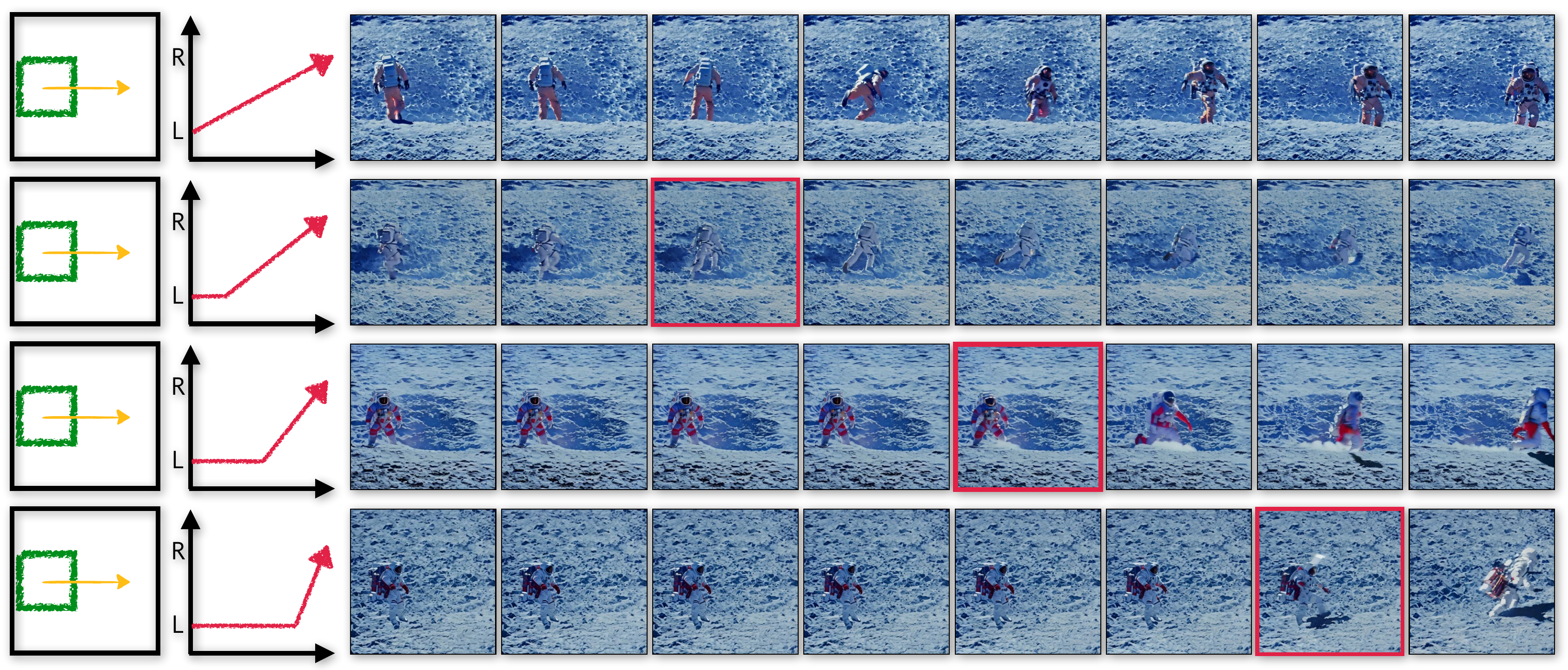}
  \caption{\textbf{Speed Test: the timing of a keyframe.} The result shows four synthesized video sequences with the subject directed according to the yellow arrow starting at the position indicated by green bbox, as illustrated in the first column. All experiments except the first use three keyframes ($|\KKK| =3$), where the timing of the internal keyframe (e.g., $\KKK_{f_1}$) controls the duration of a stationary phase and the speed of the subsequent motion,   as illustrated in the second column.
  The horizontal and vertical axis in the second column represent the left/right position and timing, respectively. The frame outlined in red indicates the frame controlled by $\KKK_{f_1}$, corresponding to the time when the astronaut starts to move. The prompt used for all experiments: ``The \textbf{astronaut} walking on the moon''.}
  \label{fig:supp_speed_key}
\end{figure*}

\FigRef{fig:supp_speed_key} shows the results of moving the subject with increasing speeds. The first row shows the astronaut moving with constant speed obtained by the linearly interpolating bboxes at the left and right of the image.  Starting from the second row, the astronaut holds the position of the first bbox on the left side of the image for some period of time, then moves more rapidly to the right side of the image, as illustrated in the second column of the figure. This is obtained by changing the timing of a single ``middle'' keyframe $\KKK_{f_1}$, where the first keyframe and the middle keyframe have the same bbox location (e.g., $\BBB_{f_0} \equiv \BBB_{f_1}$).
 Similar to the results in \FigRef{fig:supp_mul_keys}, the synthesis may generate motion blur and artifacts when the speed is high (e.g., last row).


\newpage
\subsection{Exploration and Ablation: Irregular trajectory}

\begin{figure*}[!htb]
  \centering
  \includegraphics[width=1.0\textwidth]{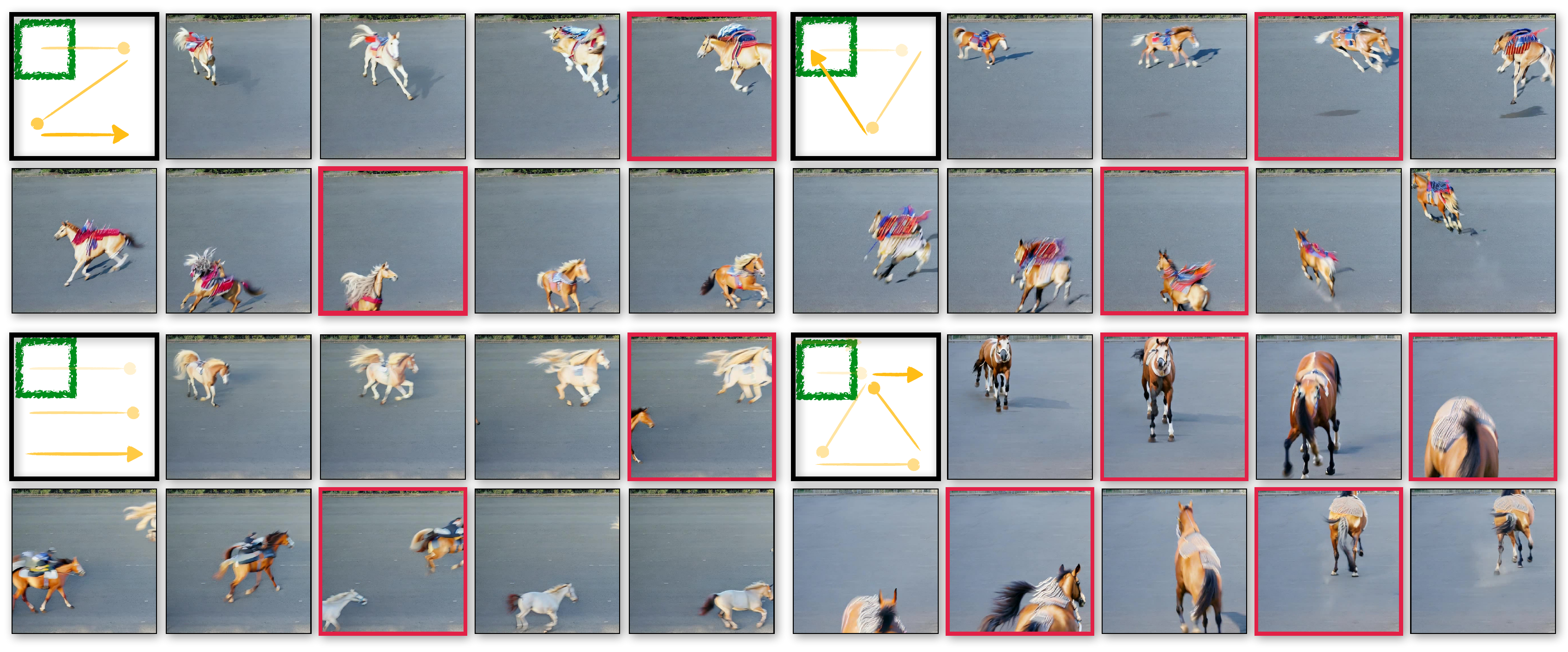}
  \caption{\textbf{Irregular trajectory.} The figure shows four synthesized video sequences with the horse subject directed according to the yellow arrows starting from the position indicated by green bbox.
  The frames highlighted in red correspond to keyframes. The start and end keyframes are not indicated. The prompt used for all examples: ``A \textbf{horse} galloping on the road''.}
  \label{fig:supp_irr_path}
\end{figure*}

We illustrate irregular trajectories determined by varied keyframes in \FigRef{fig:supp_irr_path}. The four experiments involve a zigzag trajectory (top-left), a triangle trajectory (top-right), a \emph{discontinuous} trajectory (bottom-left), and a down-pointing triangle trajectory (bottom-right). In every result the horse shows high-speed running with motion blur. However, the results with turning points show limitations in depicting the horse quickly turning around and may show artifacts. For example, in the third frame of the down-pointing triangle case, the horse appears to swap its head and tail. Difficulty portraying this turn is somewhat expected, as horses cannot naturally execute tight high-speed turns, unlike cats or dogs. On the other hand, the down-pointing triangle video naturally introduces a perspective-like size change as the horse moves higher in the image,
similar to the previous results in \FigRef{fig:supp_dyn_bbox}, and also the tiger example \FigRef{fig:main_dyn} in our main text. In summary, maintaining consistency between the prompt and the timing and location of the keyframed bounding boxes is crucial for producing realistic results.

\small
\bibliographystyle{ieeenat_fullname}
\bibliography{main}

\begin{thebibliography}{55}
\providecommand{\natexlab}[1]{#1}
\providecommand{\url}[1]{\texttt{#1}}
\expandafter\ifx\csname urlstyle\endcsname\relax
  \providecommand{\doi}[1]{doi: #1}\else
  \providecommand{\doi}{doi: \begingroup \urlstyle{rm}\Url}\fi

\bibitem[Arijon(1976)]{FilmGrammar}
Daniel Arijon.
\newblock \emph{Grammar of the Film Language}.
\newblock Focal Press, 1976.

\bibitem[Balaji et~al.(2022)Balaji, Nah, Huang, Vahdat, Song, Kreis, Aittala, Aila, Laine, Catanzaro, Karras, and Liu]{ediffI}
Yogesh Balaji, Seungjun Nah, Xun Huang, Arash Vahdat, Jiaming Song, Karsten Kreis, Miika Aittala, Timo Aila, Samuli Laine, Bryan Catanzaro, Tero Karras, and Ming{-}Yu Liu.
\newblock ediff-i: Text-to-image diffusion models with an ensemble of expert denoisers.
\newblock \emph{CoRR}, abs/2211.01324, 2022.

\bibitem[Bar{-}Tal et~al.(2023)Bar{-}Tal, Yariv, Lipman, and Dekel]{multidiffusion}
Omer Bar{-}Tal, Lior Yariv, Yaron Lipman, and Tali Dekel.
\newblock Multidiffusion: Fusing diffusion paths for controlled image generation.
\newblock \emph{CoRR}, abs/2302.08113, 2023.

\bibitem[Blattmann et~al.(2023)Blattmann, Rombach, Ling, Dockhorn, Kim, Fidler, and Kreis]{blattmann2023videoldm}
Andreas Blattmann, Robin Rombach, Huan Ling, Tim Dockhorn, Seung~Wook Kim, Sanja Fidler, and Karsten Kreis.
\newblock Align your latents: High-resolution video synthesis with latent diffusion models.
\newblock In \emph{IEEE Conference on Computer Vision and Pattern Recognition ({CVPR})}, 2023.

\bibitem[cerspense(2023)]{zeroscope}
cerspense.
\newblock zeroscope-v2-576w, 2023.
\newblock Accessed: 2023-10-01.

\bibitem[Chen et~al.(2023)Chen, Wu, Xie, Wu, Li, Xia, Xiao, and Lin]{chen2023controlavideo}
Weifeng Chen, Jie Wu, Pan Xie, Hefeng Wu, Jiashi Li, Xin Xia, Xuefeng Xiao, and Liang Lin.
\newblock Control-a-video: Controllable text-to-video generation with diffusion models, 2023.

\bibitem[Esser et~al.(2023)Esser, Chiu, Atighehchian, Granskog, and Germanidis]{Esser2023StructureAC}
Patrick Esser, Johnathan Chiu, Parmida Atighehchian, Jonathan Granskog, and Anastasis Germanidis.
\newblock Structure and content-guided video synthesis with diffusion models.
\newblock \emph{ArXiv}, abs/2302.03011, 2023.

\bibitem[Ge et~al.(2023)Ge, Nah, Liu, Poon, Tao, Catanzaro, Jacobs, Huang, Liu, and Balaji]{nvidianoiseprior}
Songwei Ge, Seungjun Nah, Guilin Liu, Tyler Poon, Andrew Tao, Bryan Catanzaro, David Jacobs, Jia-Bin Huang, Ming-Yu Liu, and Yogesh Balaji.
\newblock Preserve your own correlation: A noise prior for video diffusion models.
\newblock \emph{Proceedings of the IEEE/CVF International Conference on Computer Vision 2023}, 2023.

\bibitem[Harvey et~al.(2022)Harvey, Naderiparizi, Masrani, Weilbach, and Wood]{harvey2022flexible}
William Harvey, Saeid Naderiparizi, Vaden Masrani, Christian Weilbach, and Frank Wood.
\newblock Flexible diffusion modeling of long videos, 2022.

\bibitem[Hertz et~al.(2022)Hertz, Mokady, Tenenbaum, Aberman, Pritch, and Cohen-Or]{PromptToPrompt}
Amir Hertz, Ron Mokady, Jay Tenenbaum, Kfir Aberman, Yael Pritch, and Daniel Cohen-Or.
\newblock Prompt-to-prompt image editing with cross attention control.
\newblock \emph{arXiv preprint arXiv:2208.01626}, 2022.

\bibitem[Heusel et~al.(2017)Heusel, Ramsauer, Unterthiner, Nessler, and Hochreiter]{FID}
Martin Heusel, Hubert Ramsauer, Thomas Unterthiner, Bernhard Nessler, and Sepp Hochreiter.
\newblock Gans trained by a two time-scale update rule converge to a local nash equilibrium.
\newblock In \emph{Advances in Neural Information Processing Systems}. Curran Associates, Inc., 2017.

\bibitem[Ho et~al.(2020)Ho, Jain, and Abbeel]{hoDDPM20}
Jonathan Ho, Ajay Jain, and Pieter Abbeel.
\newblock Denoising diffusion probabilistic models.
\newblock \emph{Advances in Neural Information Processing Systems}, 33, 2020.

\bibitem[Ho et~al.(2022{\natexlab{a}})Ho, Chan, Saharia, Whang, Gao, Gritsenko, Kingma, Poole, Norouzi, Fleet, and Salimans]{Ho2022ImagenVH}
Jonathan Ho, William Chan, Chitwan Saharia, Jay Whang, Ruiqi Gao, Alexey~A. Gritsenko, Diederik~P. Kingma, Ben Poole, Mohammad Norouzi, David~J. Fleet, and Tim Salimans.
\newblock Imagen video: High definition video generation with diffusion models.
\newblock \emph{ArXiv}, abs/2210.02303, 2022{\natexlab{a}}.

\bibitem[Ho et~al.(2022{\natexlab{b}})Ho, Salimans, Gritsenko, Chan, Norouzi, and Fleet]{ho2022video}
Jonathan Ho, Tim Salimans, Alexey Gritsenko, William Chan, Mohammad Norouzi, and David~J. Fleet.
\newblock Video diffusion models, 2022{\natexlab{b}}.

\bibitem[Hu and Xu(2023)]{hu2023videocontrolnet}
Zhihao Hu and Dong Xu.
\newblock Videocontrolnet: A motion-guided video-to-video translation framework by using diffusion model with controlnet, 2023.

\bibitem[Huggingface(2023)]{huggingface}
Huggingface.
\newblock Stable diffusion 1 demo, 2023.
\newblock Accessed: 2023-01-01.

\bibitem[Höppe et~al.(2022)Höppe, Mehrjou, Bauer, Nielsen, and Dittadi]{Infilling2022}
Tobias Höppe, Arash Mehrjou, Stefan Bauer, Didrik Nielsen, and Andrea Dittadi.
\newblock Diffusion models for video prediction and infilling, 2022.

\bibitem[Jain et~al.(2023)Jain, Nasery, Vineet, and Behl]{jain2023peekaboo}
Yash Jain, Anshul Nasery, Vibhav Vineet, and Harkirat Behl.
\newblock Peekaboo: Interactive video generation via masked-diffusion, 2023.

\bibitem[Khachatryan et~al.(2023)Khachatryan, Movsisyan, Tadevosyan, Henschel, Wang, Navasardyan, and Shi]{text2video-zero}
Levon Khachatryan, Andranik Movsisyan, Vahram Tadevosyan, Roberto Henschel, Zhangyang Wang, Shant Navasardyan, and Humphrey Shi.
\newblock Text2video-zero: Text-to-image diffusion models are zero-shot video generators.
\newblock \emph{arXiv preprint arXiv:2303.13439}, 2023.

\bibitem[Li et~al.(2023)Li, Liu, Wu, Mu, Yang, Gao, Li, and Lee]{GLIGEN}
Yuheng Li, Haotian Liu, Qingyang Wu, Fangzhou Mu, Jianwei Yang, Jianfeng Gao, Chunyuan Li, and Yong~Jae Lee.
\newblock {GLIGEN:} open-set grounded text-to-image generation.
\newblock \emph{CoRR}, abs/2301.07093, 2023.

\bibitem[Lian et~al.(2023)Lian, Shi, Yala, Darrell, and Li]{lian2023llmgrounded}
Long Lian, Baifeng Shi, Adam Yala, Trevor Darrell, and Boyi Li.
\newblock Llm-grounded video diffusion models, 2023.

\bibitem[Liew et~al.(2022)Liew, Yan, Zhou, and Feng]{Magicmix}
Jun~Hao Liew, Hanshu Yan, Daquan Zhou, and Jiashi Feng.
\newblock Magicmix: Semantic mixing with diffusion models.
\newblock \emph{CoRR}, abs/2210.16056, 2022.

\bibitem[Luo et~al.(2023)Luo, Chen, Zhang, Huang, Wang, Shen, Zhao, Zhou, and Tan]{luo2023videofusion}
Zhengxiong Luo, Dayou Chen, Yingya Zhang, Yan Huang, Liang Wang, Yujun Shen, Deli Zhao, Jingren Zhou, and Tieniu Tan.
\newblock Videofusion: Decomposed diffusion models for high-quality video generation, 2023.

\bibitem[Ma et~al.(2023)Ma, Lewis, Lahiri, Leung, and Kleijn]{ma2023directed}
Wan-Duo~Kurt Ma, J.~P. Lewis, Avisek Lahiri, Thomas Leung, and W.~Bastiaan Kleijn.
\newblock Directed diffusion: Direct control of object placement through attention guidance, 2023.

\bibitem[Minderer et~al.(2022)Minderer, Gritsenko, Stone, Neumann, Weissenborn, Dosovitskiy, Mahendran, Arnab, Dehghani, Shen, Wang, Zhai, Kipf, and Houlsby]{OWLViTlarge}
Matthias Minderer, Alexey Gritsenko, Austin Stone, Maxim Neumann, Dirk Weissenborn, Alexey Dosovitskiy, Aravindh Mahendran, Anurag Arnab, Mostafa Dehghani, Zhuoran Shen, Xiao Wang, Xiaohua Zhai, Thomas Kipf, and Neil Houlsby.
\newblock Simple open-vocabulary object detection.
\newblock In \emph{Computer Vision -- ECCV 2022}, pages 728--755, Cham, 2022. Springer Nature Switzerland.

\bibitem[Mou et~al.(2023)Mou, Wang, Xie, Wu, Zhang, Qi, Shan, and Qie]{mou2023t2i}
Chong Mou, Xintao Wang, Liangbin Xie, Yanze Wu, Jian Zhang, Zhongang Qi, Ying Shan, and Xiaohu Qie.
\newblock T2i-adapter: Learning adapters to dig out more controllable ability for text-to-image diffusion models.
\newblock \emph{arXiv preprint arXiv:2302.08453}, 2023.

\bibitem[Ng et~al.(2022)Ng, Ong, Zheng, Ni, Yeo, and Liu]{Ng_2022_CVPR}
Xun~Long Ng, Kian~Eng Ong, Qichen Zheng, Yun Ni, Si~Yong Yeo, and Jun Liu.
\newblock Animal kingdom: A large and diverse dataset for animal behavior understanding.
\newblock In \emph{Proceedings of the IEEE/CVF Conference on Computer Vision and Pattern Recognition (CVPR)}, pages 19023--19034, 2022.

\bibitem[Nichol and Dhariwal(2021)]{nichol2021improved}
Alex Nichol and Prafulla Dhariwal.
\newblock Improved denoising diffusion probabilistic models, 2021.

\bibitem[Nichol et~al.(2022)Nichol, Dhariwal, Ramesh, Shyam, Mishkin, McGrew, Sutskever, and Chen]{GLIDE}
Alexander~Quinn Nichol, Prafulla Dhariwal, Aditya Ramesh, Pranav Shyam, Pamela Mishkin, Bob McGrew, Ilya Sutskever, and Mark Chen.
\newblock {GLIDE:} towards photorealistic image generation and editing with text-guided diffusion models.
\newblock In \emph{ICML}, 2022.

\bibitem[Paszke et~al.(2019)Paszke, Gross, Massa, Lerer, Bradbury, Chanan, Killeen, Lin, Gimelshein, Antiga, Desmaison, Köpf, Yang, DeVito, Raison, Tejani, Chilamkurthy, Steiner, Fang, Bai, and Chintala]{paszke2019pytorch}
Adam Paszke, Sam Gross, Francisco Massa, Adam Lerer, James Bradbury, Gregory Chanan, Trevor Killeen, Zeming Lin, Natalia Gimelshein, Luca Antiga, Alban Desmaison, Andreas Köpf, Edward Yang, Zach DeVito, Martin Raison, Alykhan Tejani, Sasank Chilamkurthy, Benoit Steiner, Lu Fang, Junjie Bai, and Soumith Chintala.
\newblock Pytorch: An imperative style, high-performance deep learning library, 2019.

\bibitem[Qi et~al.(2023)Qi, Cun, Zhang, Lei, Wang, Shan, and Chen]{qi2023fatezero}
Chenyang Qi, Xiaodong Cun, Yong Zhang, Chenyang Lei, Xintao Wang, Ying Shan, and Qifeng Chen.
\newblock Fatezero: Fusing attentions for zero-shot text-based video editing.
\newblock \emph{arXiv:2303.09535}, 2023.

\bibitem[Radford et~al.(2021)Radford, Kim, Hallacy, Ramesh, Goh, Agarwal, Sastry, Askell, Mishkin, Clark, Krueger, and Sutskever]{CLIP}
Alec Radford, Jong~Wook Kim, Chris Hallacy, Aditya Ramesh, Gabriel Goh, Sandhini Agarwal, Girish Sastry, Amanda Askell, Pamela Mishkin, Jack Clark, Gretchen Krueger, and Ilya Sutskever.
\newblock Learning transferable visual models from natural language supervision.
\newblock In \emph{Proc.~ICML}, 2021.

\bibitem[Ramesh et~al.(2022)Ramesh, Dhariwal, Nichol, Chu, and Chen]{DALLE2}
Aditya Ramesh, Prafulla Dhariwal, Alex Nichol, Casey Chu, and Mark Chen.
\newblock Hierarchical text-conditional image generation with {CLIP} latents.
\newblock \emph{CoRR}, abs/2204.06125, 2022.

\bibitem[Rombach et~al.(2022)Rombach, Blattmann, Lorenz, Esser, and Ommer]{RombachStablediffusion21}
Robin Rombach, Andreas Blattmann, Dominik Lorenz, Patrick Esser, and Bj\"orn Ommer.
\newblock High-resolution image synthesis with latent diffusion models.
\newblock In \emph{Proceedings of the IEEE/CVF Conference on Computer Vision and Pattern Recognition (CVPR)}, pages 10684--10695, 2022.

\bibitem[Saharia et~al.(2022{\natexlab{a}})Saharia, Chan, Saxena, Li, Whang, Denton, Ghasemipour, Ayan, Mahdavi, Lopes, Salimans, Ho, Fleet, and Norouzi]{IMAGEN}
Chitwan Saharia, William Chan, Saurabh Saxena, Lala Li, Jay Whang, Emily Denton, Seyed Kamyar~Seyed Ghasemipour, Burcu~Karagol Ayan, S.~Sara Mahdavi, Rapha~Gontijo Lopes, Tim Salimans, Jonathan Ho, David~J. Fleet, and Mohammad Norouzi.
\newblock Photorealistic text-to-image diffusion models with deep language understanding.
\newblock \emph{CoRR}, abs/2205.11487, 2022{\natexlab{a}}.

\bibitem[Saharia et~al.(2022{\natexlab{b}})Saharia, Chan, Saxena, Li, Whang, Denton, Ghasemipour, Ayan, Mahdavi, Lopes, Salimans, Ho, Fleet, and Norouzi]{Saharia2022PhotorealisticTD}
Chitwan Saharia, William Chan, Saurabh Saxena, Lala Li, Jay Whang, Emily~L. Denton, Seyed Kamyar~Seyed Ghasemipour, Burcu~Karagol Ayan, Seyedeh~Sara Mahdavi, Raphael~Gontijo Lopes, Tim Salimans, Jonathan Ho, David~J. Fleet, and Mohammad Norouzi.
\newblock Photorealistic text-to-image diffusion models with deep language understanding.
\newblock \emph{ArXiv}, abs/2205.11487, 2022{\natexlab{b}}.

\bibitem[Sohl-Dickstein et~al.(2015)Sohl-Dickstein, Weiss, Maheswaranathan, and Ganguli]{sohldickstein15}
Jascha Sohl-Dickstein, Eric Weiss, Niru Maheswaranathan, and Surya Ganguli.
\newblock Deep unsupervised learning using nonequilibrium thermodynamics.
\newblock In \emph{International Conference on Machine Learning}, 2015.

\bibitem[Song et~al.(2021)Song, Meng, and Ermon]{DDIM}
Jiaming Song, Chenlin Meng, and Stefano Ermon.
\newblock Denoising diffusion implicit models.
\newblock In \emph{9th International Conference on Learning Representations, {ICLR} 2021, Virtual Event, Austria, May 3-7, 2021}, 2021.

\bibitem[Song and Ermon(2019)]{songScorematching19}
Yang Song and Stefano Ermon.
\newblock Generative modeling by estimating gradients of the data distribution.
\newblock In \emph{NeurIPS}, 2019.

\bibitem[Sun and Wu(2022)]{L2Igan}
Wei Sun and Tianfu Wu.
\newblock Learning layout and style reconfigurable gans for controllable image synthesis.
\newblock \emph{TPAMI}, 44:\penalty0 5070--5087, 2022.

\bibitem[Szab{\'{o}} and Horv{\'{a}}th(2021)]{szabo-centeredobjects-21}
Gergely Szab{\'{o}} and Andr{\'{a}}s Horv{\'{a}}th.
\newblock Mitigating the bias of centered objects in common datasets.
\newblock \emph{CoRR}, abs/2112.09195, 2021.

\bibitem[Vaswani et~al.(2017)Vaswani, Shazeer, Parmar, Uszkoreit, Jones, Gomez, Kaiser, and Polosukhin]{allyouneedattn}
Ashish Vaswani, Noam Shazeer, Niki Parmar, Jakob Uszkoreit, Llion Jones, Aidan~N Gomez, \L~ukasz Kaiser, and Illia Polosukhin.
\newblock Attention is all you need.
\newblock In \emph{Advances in Neural Information Processing Systems}. Curran Associates, Inc., 2017.

\bibitem[Voleti et~al.(2022)Voleti, Jolicoeur-Martineau, and Pal]{voleti2022MCVD}
Vikram Voleti, Alexia Jolicoeur-Martineau, and Christopher Pal.
\newblock Mcvd: Masked conditional video diffusion for prediction, generation, and interpolation.
\newblock In \emph{(NeurIPS) Advances in Neural Information Processing Systems}, 2022.

\bibitem[Wang et~al.(2023)Wang, Yuan, Chen, Zhang, Wang, and Zhang]{modelscope}
Jiuniu Wang, Hangjie Yuan, Dayou Chen, Yingya Zhang, Xiang Wang, and Shiwei Zhang.
\newblock Modelscope text-to-video technical report, 2023.

\bibitem[Wang et~al.(2024)Wang, Zhang, Zou, Zeng, Wei, Yuan, and Li]{wang2024boximator}
Jiawei Wang, Yuchen Zhang, Jiaxin Zou, Yan Zeng, Guoqiang Wei, Liping Yuan, and Hang Li.
\newblock Boximator: Generating rich and controllable motions for video synthesis, 2024.

\bibitem[Weng(2021)]{blogLillog}
Lilian Weng.
\newblock What are diffusion models?, 2021.

\bibitem[wiki(2023)]{keyframe}
wiki.
\newblock keyframe, 2023.
\newblock Accessed: 2023-10-01.

\bibitem[Wu et~al.(2023)Wu, Ge, Wang, Lei, Gu, Shi, Hsu, Shan, Qie, and Shou]{wu2023tune}
Jay~Zhangjie Wu, Yixiao Ge, Xintao Wang, Stan~Weixian Lei, Yuchao Gu, Yufei Shi, Wynne Hsu, Ying Shan, Xiaohu Qie, and Mike~Zheng Shou.
\newblock Tune-a-video: One-shot tuning of image diffusion models for text-to-video generation.
\newblock In \emph{Proceedings of the IEEE/CVF International Conference on Computer Vision}, pages 7623--7633, 2023.

\bibitem[Xie et~al.(2023)Xie, Li, Huang, Liu, Zhang, Zheng, and Shou]{boxdiff}
Jinheng Xie, Yuexiang Li, Yawen Huang, Haozhe Liu, Wentian Zhang, Yefeng Zheng, and Mike~Zheng Shou.
\newblock Boxdiff: Text-to-image synthesis with training-free box-constrained diffusion.
\newblock \emph{CoRR}, abs/2307.10816, 2023.

\bibitem[Yan et~al.(2023)Yan, Liew, Mai, Lin, and Feng]{yan2023magicprop}
Hanshu Yan, Jun~Hao Liew, Long Mai, Shanchuan Lin, and Jiashi Feng.
\newblock Magicprop: Diffusion-based video editing via motion-aware appearance propagation, 2023.

\bibitem[Yang et~al.(2022{\natexlab{a}})Yang, Srivastava, and Mandt]{yang2022diffusion}
Ruihan Yang, Prakhar Srivastava, and Stephan Mandt.
\newblock Diffusion probabilistic modeling for video generation, 2022{\natexlab{a}}.

\bibitem[Yang et~al.(2024)Yang, Hou, Huang, Ma, Wan, Zhang, Chen, and Liao]{yang2024directavideo}
Shiyuan Yang, Liang Hou, Haibin Huang, Chongyang Ma, Pengfei Wan, Di Zhang, Xiaodong Chen, and Jing Liao.
\newblock Direct-a-video: Customized video generation with user-directed camera movement and object motion, 2024.

\bibitem[Yang et~al.(2022{\natexlab{b}})Yang, Liu, Wang, Yang, and Tao]{L2Ixformer}
Zuopeng Yang, Daqing Liu, Chaoyue Wang, J. Yang, and Dacheng Tao.
\newblock Modeling image composition for complex scene generation.
\newblock \emph{CVPR}, pages 7754--7763, 2022{\natexlab{b}}.

\bibitem[Zhang and Agrawala(2023)]{controlnet}
Lvmin Zhang and Maneesh Agrawala.
\newblock Adding conditional control to text-to-image diffusion models, 2023.

\bibitem[Zhao et~al.(2020)Zhao, Yin, Meng, and Sigal]{layout2im}
Bo Zhao, Weidong Yin, Lili Meng, and Leonid Sigal.
\newblock Layout2image: Image generation from layout.
\newblock \emph{Int. J. Comput. Vis.}, 128\penalty0 (10):\penalty0 2418--2435, 2020.

\end{thebibliography}
\end{document}